%% file: main.tex
\documentclass[letterpaper]{article} 
\usepackage{aaai2026}  
\usepackage{times}  
\usepackage{helvet}  
\usepackage{courier}  
\usepackage[hyphens]{url}  
\usepackage{graphicx} 
\usepackage{natbib}  
\usepackage{caption} 
\usepackage{algorithm}

\usepackage{newfloat}
\usepackage{listings}
\DeclareCaptionStyle{ruled}{labelfont=normalfont,labelsep=colon,strut=off} 
\floatstyle{ruled}
\newfloat{listing}{tb}{lst}{}
\floatname{listing}{Listing}
\input{custom_imports}

\nocopyright

\title{Graph-of-Mark: Promote Spatial Reasoning in Multimodal Language Models with Graph-Based Visual Prompting}

\author{
    Giacomo Frisoni\equalcontrib, Lorenzo Molfetta\equalcontrib, Mattia Buzzoni\equalcontrib, Gianluca Moro\equalcontrib
}

\affiliations{
    Department of Computer Science and Engineering, University of Bologna \\
    \{giacomo.frisoni, lorenzo.molfetta, gianluca.moro\}@unibo.it,\\ mattia.buzzoni@studio.unibo.it
}

\begin{document}

\maketitle


\begin{abstract}
Recent advances in training-free visual prompting, such as Set-of-Mark, have emerged as a promising direction for enhancing the grounding capabilities of multimodal language models (MLMs). These techniques operate by partitioning the input image into object regions and annotating them with marks--predominantly boxes with numeric identifiers--before feeding the augmented image to the MLM. However, these approaches treat marked objects as isolated entities, failing to capture the \textit{relationships} between them. On these premises, we propose Graph-of-Mark (\gom), the first pixel-level visual prompting technique that overlays scene graphs onto the input image for spatial reasoning tasks. We evaluate \gom across 3 open-source MLMs and 4 different datasets, conducting extensive ablations on drawn components and investigating the impact of auxiliary graph descriptions in the text prompt. Our results demonstrate that \gom consistently improves the zero-shot capability of MLMs in interpreting object positions and relative directions, improving base accuracy in visual question answering and localization up to 11 percentage points.\footnote{The definitive, copyrighted, and peer-reviewed version of this article is published in \textit{AAAI 2026}, edited by Sven Koenig et al., AAAI Press, Vol. 40, No. 36, Technical Track, pp. 30726-30734, 2026. DOI: \url{https://doi.org/10.1609/aaai.v40i36.40329}.}
\end{abstract}
\begin{links}
    \link{\gom}{https://github.com/disi-unibo-nlp/graph-of-marks}
\end{links}


\section{Introduction}

It takes more than detecting objects to make sense of a visual scene.
Often, it is the \textit{arrangement} of objects in space that reveals the underlying semantics of complex 2D and 3D environments.
Despite the proliferation and remarkable progress of multimodal language models (MLMs)~\cite{Yin2024ASO}, spatial reasoning remains a challenge in machine perception~\cite{DBLP:conf/emnlp/KamathHC23a}.
Recent empirical studies show that even state-of-the-art models tend to overlook positional aspects and view images as mere \quotes{bags of objects}~\cite{DBLP:conf/emnlp/HerzigMKAFDG23,DBLP:conf/cvpr/DovehAHSHGFPUK23}.
This limitation is deeply rooted in their architecture and training objectives, which prioritize global image-level understanding over explicit spatial supervision with localized representations and compositionality~\cite{DBLP:journals/corr/abs-2504-15037}.
As a result, MLMs exhibit persistent difficulty in distinguishing even basic spatial concepts~\cite{DBLP:conf/eccv/FuHLFWLRSMK24,DBLP:conf/cvpr/MajumdarA0PYHSM24,DBLP:conf/acl/ZhangCXL24,DBLP:conf/emnlp/ShiriGF0HL24} such as \textit{left} and \textit{right}, \textit{above} and \textit{below}, \textit{near}, and \textit{behind}--the latter requiring depth cues not captured by standard RGB pixel data.
Yet, spatial reasoning is far from an academic abstraction--it is fundamental to real-world applications, spanning fields from biomedical research and clinical practice~\cite{DBLP:conf/ic3k/DomeniconiMMP14,DBLP:journals/bmcbi/LenaDMM15} to GUI agents, augmented reality, robotic manipulation, autonomous navigation and neuro-symbolic systems~\cite{DBLP:conf/ijcai/DelvecchioMM25}.

\begin{figure}[t]
    \centering
    \includegraphics[width=\linewidth]{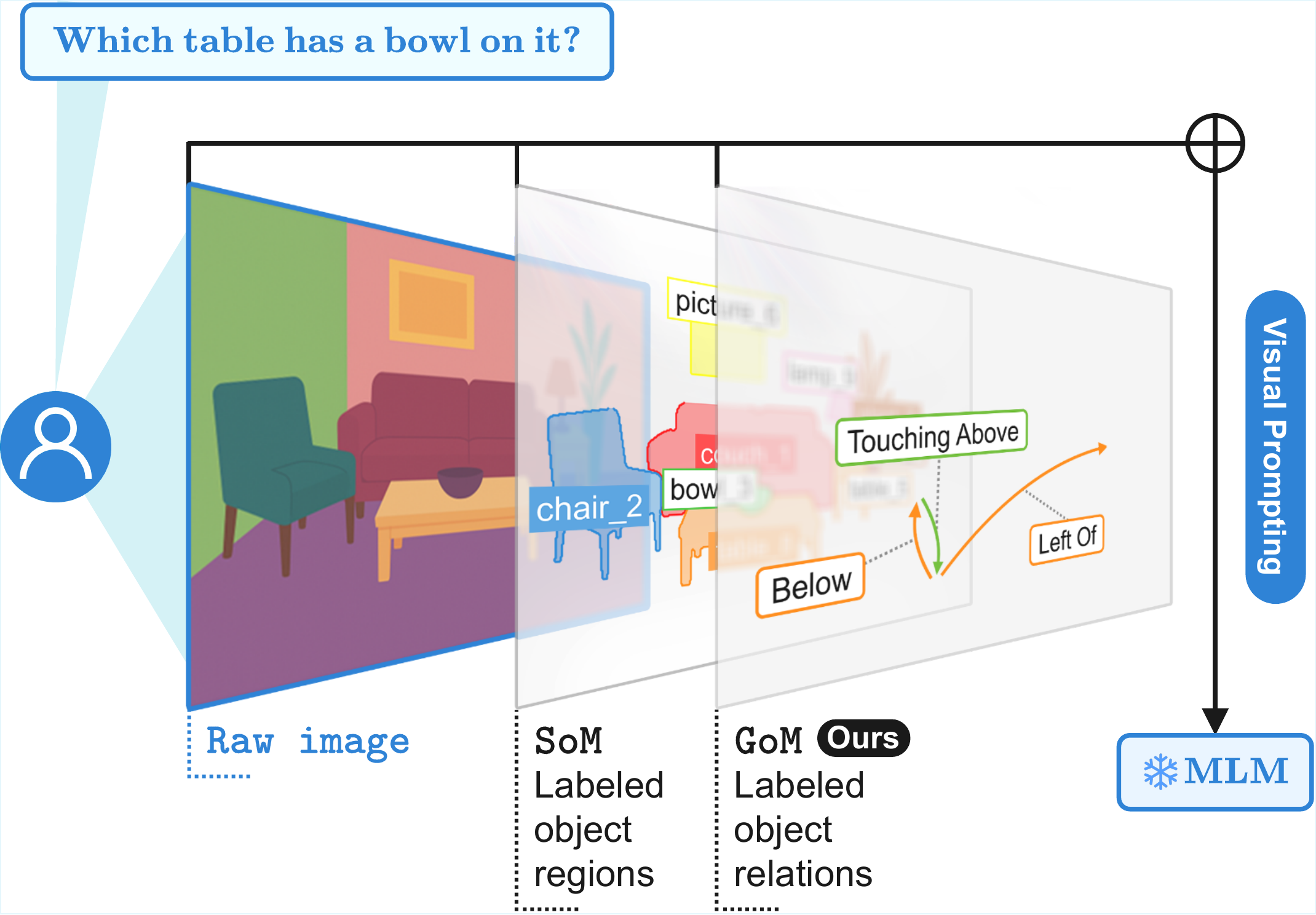}
    \caption{\textbf{Illustration of \gom.} A multimodal language model is prompted by anchoring the input image in scene graphs expressing spatial object relations that are relevant to solving the task query provided by the user.}
    \label{fig:gom-abstract}
\end{figure}

A natural response to these limitations has been to inject spatial awareness into MLMs by fine-tuning~\cite{DBLP:conf/cvpr/0003XKISGX24,DBLP:journals/corr/abs-2506-04220}.
However, this route is computationally expensive and inflexible as the models must be retrained to handle new tasks and domains, a challenge often observed in cross-domain generalization~\cite{DBLP:conf/ic3k/DomeniconiMPS14,DBLP:conf/ic3k/MoroPP018}.
An increasingly attractive alternative has emerged in the form of prompting, where task-specific guidance is provided at inference time through carefully crafted inputs.
Since textual prompts inherently struggle to convey fine-grained or dense information from image modalities~\cite{DBLP:journals/corr/abs-2407-04681}, \textit{visual prompting} offers a more effective mechanism to uncover the spatial reasoning potential hidden within MLMs.
Existing methods can be divided into two classes: \textit{embedding-level (soft)}, which encodes visual items into latent features, and \textit{pixel-level (hard)}, which renders visible marks--such as masks, boxes, arrows, numbers, or alphabets--onto the image.
Among the latter, Set-of-Mark (SoM)~\cite{DBLP:journals/corr/abs-2310-11441} has rapidly established itself as a de facto standard for improving visual grounding by overlaying numbered regions.
Its effectiveness has inspired a growing line of follow-up work, including OmniParser~\cite{DBLP:journals/corr/abs-2408-00203}.
Despite these advances, marked objects are predominantly treated as independent entities--overlooking the relational structure that governs scenes and leaving a rich modeling layer still untapped.

In this paper, we introduce Graph-of-Mark (\gom), the first training-free, pixel-level visual prompting technique that embeds graph-connected marks for zero-shot spatial inference.
Before feeding the model, \gom draws a \textit{scene graph} (SG) over the raw input image--nodes correspond to detected object instances, while edges denote their spatial relations (Figure~\ref{fig:gom-abstract}).
Designed as a lightweight \textit{plug-and-play} module compatible with any MLMs, \gom constructs SGs automatically and at scale using open-vocabulary object detectors, segmenters, and depth estimators.
Moreover, \gom is broadly applicable, as the underlying SGs demand no ground-truth annotations and are both task- and domain-agnostic.

We validate the effectiveness of \gom through rigorous quantitative evaluation using 3 open-source MLMs and 4 heterogeneous, publicly available datasets spanning 2 vision tasks: visual question answering (VQA) and referring expression comprehension (REC).
Our results demonstrate that \gom consistently enhances the zero-shot capabilities of all tested models, surpassing the performance of the existing most popular image prompting techniques. 
We further conduct ablation studies to dissect the contribution of edge labels beyond simple connectivity and the relative effectiveness of numeric versus alphabetic labels.
Finally, we examine the impact of auxiliary SG descriptions included in the text prompt, probing whether MLMs can internalize the relational structure from visual input alone.
To support reproducibility and foster further innovation, we release \gom--code, preprocessed dataset images, and evaluation scripts--under an open-source MIT license.
We hope that \gom will provide fertile ground for the development of spatial-aware MLMs and hybrid graph-language solutions.


\section{Related Work}

\paragraph{Multimodal Language Models for Spatial Reasoning.}

The advancement of MLMs has enabled unified reasoning over text and vision inputs within scalable general-purpose architectures~\cite{DBLP:conf/nips/ChenLDZZCDWQLZ24}.
Still, frontier models fall short in tasks involving visual evidence grounding~\cite{DBLP:conf/cvpr/XiaoY0C24} and spatial intelligence \cite{DBLP:conf/cvpr/YangYGH0X25,DBLP:conf/iclr/RamakrishnanWKK25}.
Models such as CLIP~\cite{DBLP:conf/icml/RadfordKHRGASAM21}, SPHINX~\cite{DBLP:journals/corr/abs-2311-07575}, LLaVA~\cite{DBLP:conf/nips/LiuLWL23a}, VisionLLM~\cite{DBLP:conf/nips/WangCCWZZLLZQD23,DBLP:conf/nips/WuZXL00WZLL0QD24}, and CogVLM2~\cite{DBLP:conf/nips/WangLYHQWJYZSXC24}, integrate visual and textual encoded representations using instruction tuning or contrastive pretraining to align modalities.
Spatial information in these systems is typically encoded as plain text coordinates~\cite{DBLP:journals/corr/abs-2306-14824} or serialized region descriptions~\cite{DBLP:journals/corr/abs-2310-09478}, placing the burden of geometric interpretation on the language backbone.
Spatial reasoning is treated more directly in models that encode structural information at the region or scene level.
SpatialVLM~\cite{DBLP:conf/cvpr/0003XKISGX24} uses the supervision of synthetic 3D question-answer pairs to model object configurations, whereas Struct2D~\cite{DBLP:journals/corr/abs-2506-04220} applies structured 2D multimodal input to guide 3D perception.
Shikra~\cite{DBLP:journals/corr/abs-2306-15195} embeds coordinate references as positional tokens within the language stream, preserving compatibility with frozen decoders. 
ASM~\cite{DBLP:conf/iclr/Wang0LWHXCLZ0CL24} leverages densely annotated region-level supervision to improve spatial grounding, while SPHINX-V~\cite{DBLP:conf/iclr/LinWA0ZL0Z025} directly encodes visual tags into the input image, enabling localized understanding with frozen LLMs.
GLaMM~\cite{DBLP:conf/cvpr/Rasheed0MS0CAX024} generates natural language responses while simultaneously providing the associated object segmentation masks, and Omni-RGPT~\cite{DBLP:conf/cvpr/HeoCHLRKWH25} introduces Token Mark to achieve consistent region representation for both images and videos by embedding tokens directly into visual feature space and text prompts. 
VISPROG~\cite{DBLP:conf/cvpr/GuptaK23} builds executable neuro-symbolic visual programs to enable spatial reasoning across different tasks.
However, these methods require expensive re-training and custom architecture, limiting their adoption in general-purpose domains. 
\gom can be applied to existing models without modification, enabling spatial reasoning with minimal overhead.
\paragraph{Image Prompting and Augmentation.}

Visual prompting has become a standard lightweight mechanism for reducing hallucinations in MLMs~\cite{DBLP:conf/nips/00020YG24}
and steering vision-language models via spatial annotations~\cite{DBLP:conf/cvpr/LiJZRLZXLYLZ024,DBLP:conf/cvpr/CaiLMMCPL24}, enabling task adaptation and spatial awareness without altering model parameters.
Early methods enhanced captioning and VQA alignment using region segmentation and bottom-up attention~\cite{DBLP:conf/cvpr/00010BT0GZ18}, or soft perturbations to improve grounding.
Prompting strategies have since incorporated more structured spatial signals.
Pixel-level modifications driven by text~\cite{DBLP:conf/eccv/YuYW24} and training-free visual prompt learning through test-time optimization ~\cite{DBLP:conf/nips/WuCJLHL0JSJ24} guide models toward relevant regions under varying prompt formulations.
Region-based architectures~\cite{DBLP:conf/cvpr/GuoMYBCYLL24,DBLP:conf/eccv/ZhangSCXSZLCL24,DBLP:conf/iclr/YouZGDZWCCY24,DBLP:journals/corr/abs-2404-07973} incorporate bounding-box descriptors for more accurate spatial alignment, while segmentation-based methods inject structured priors such as SAM-derived masks~\cite{DBLP:conf/cvpr/YuanLLTLQZZ24,DBLP:conf/iccv/KirillovMRMRGXW23} or exploit region-level contrastive prompts~\cite{DBLP:conf/eccv/WanCSB24} to sharpen spatial attention.
More recent formulations reframe prompting as symbolic supervision over discrete visual elements.
CPT~\cite{DBLP:journals/aiopen/YaoZZLCS24} encodes color-coded markers as fill-in-the-blank targets.
Set-of-Marks~\cite{DBLP:journals/corr/abs-2310-11441} overlays alphanumeric tags on segmented regions to support co-reference and alignment across modalities, while Magma~\cite{DBLP:conf/cvpr/YangTWZPLGCYJD025} enhances robot manipulation and UI navigation by introducing Trace-of-Mark to extract the evolving position of overlayed marks.
SpatialRGPT~\cite{DBLP:conf/nips/ChengYFGYK0L24} learns regional representations from 3D scene graphs and integrates depth information into the visual encoder to enhance spatial reasoning, while ROI-aware inputs support medical VQA~\cite{DBLP:journals/corr/abs-2410-20327,DBLP:conf/naacl/ZhuQYJLZL25}. 
OmniParser~\cite{DBLP:journals/corr/abs-2408-00203} treats UI images as structured visual documents, decomposing them into typed layout elements.
Although these methods achieve good results at lower costs, they do not explicitly leverage logical or positional spatial relationships, instead relying on existing models to interpret the augmented visual scene.

\paragraph{Structured Graph Scene Injection.}
Spatial layouts in vision-language tasks have been leveraged to construct scene-level graphs that encode object relations and guide attention toward regions relevant for reasoning~\cite{DBLP:conf/wacv/YangXGHCD0Z24}.
Region-level graph structures have been combined with GCNs for image captioning~\cite{DBLP:conf/eccv/YaoPLM18} and VQA~\cite{DBLP:journals/ijon/YusufFMHLD25}, and with random-walk mechanisms~\cite{DBLP:conf/iccv/0002YRWL23} to support relational queries.
Question-conditioned attention over scene graphs~\cite{DBLP:conf/iccv/LiGCL19} and contrastive learning for unsupervised graph construction~\cite{DBLP:conf/iccvw/SouzaAPR23} further enhance graph-based VQA representations.
Scene graphs have also been integrated into multimodal pipelines that address referring expression resolution~\cite{DBLP:conf/emnlp/Wu0LW23}, visual commonsense reasoning~\cite{DBLP:conf/aaai/WangYLZPLCC22}, and object-level spatial attention in transformers~\cite{DBLP:conf/eccv/KantBASPLA20}.
Several approaches enrich these graphs with external knowledge.
KRISP~\cite{DBLP:conf/cvpr/MarinoCP0R21} combines symbolic and LLM-derived features, while methods like MAIL~\cite{DBLP:conf/acl/DongZZZZH24} generate concept graphs through multi-stage prompting.
The explicit incorporation of external graph knowledge into neural models via relational attention are explored in~\cite{DBLP:conf/cvpr/ZhangJZ21}.
Recent methods integrate graph construction with structured prompting.
ConceptGraphs~\cite{DBLP:conf/icra/GuKMJSARPECGMTT24} generates 3D scene graphs with the support of an LLM and converts them into structured textual descriptions to support spatial planning, while Compositional CoT~\cite{DBLP:conf/cvpr/MitraHDH24} employs synthetic textual scene graphs to prompt LLMs for enhanced multi-step reasoning and image interpretation.
Despite these advancements, existing methods either fuse scene graphs through latent mechanisms or convert them into text, hindering direct visual grounding of structured information.
\gom, on the other hand, is the first to embed the graph directly into the image, enabling multimodal models to leverage relational cues at the pixel level without relying on implicit or textual intermediaries that obscure structure and reduce interpretability.


\section{Method}

In this section, we describe \gom prompting, designed to shift MLM perception from object collections to object networks.
A compact algorithmic overview of \gom, expressed in pseudocode, is made available in the Supplementary Material.

\subsection{Problem Definition}

We consider MLMs capable of reasoning over both vision and language input modalities.
Given an RGB image observation $I \in \mathbb{R}^{H \times W \times 3}$ and an associated text prompt $T$ comprising a task prefix and instance-specific query, these models generate a distribution $P_{\text{MLM}}(\cdot|I,T)$ over textual completions that serve as responses.
Although the exact parameterized MLM submodules (vision encoder, language encoder, and decoder) differ between models, the overarching computational flow remains consistent.
Recent studies have demonstrated that visual tokens fed into MLMs largely influence the attention map values, thus indirectly controlling the model output~\cite{DBLP:conf/nips/WuCJLHL0JSJ24}.
Building on this premise, we propose a training-free method that strategically alters visual tokens to induce a global perspective encompassing not only individual objects--the primary units for visual reasoning--but also their interactions.
To this end, we augment the input image $I$ to produce a scene graph-annotated image $I^{\text{SG}}$ that explicitly encodes spatial relationships between objects.
\begin{equation*}
P_{\text{MLM}}(\cdot|\underbrace{\gom(I)}_{I^{\text{SG}}},T)
\end{equation*}

\subsection{Graph-Based Image Augmentation}

\paragraph{Object Detection and Segmentation.}
Identifying the objects present in $I$, their classes, and positions constitutes the initial stage of the SG building.
We adopt a coarse-to-fine approach to partition $I$ into a set of regions $R = \{r_1, \ldots, r_{|R|}\}$, where $r_i$ is the spatial extent of the object $i$.\\
\noindent\textbf{\textit{Object detection}} We first determine the 2D bounding box coordinates and the class label of each object.
Unlike previous works that rely on single detection models, we harness an \textit{ensemble} of independent detectors with complementary strengths to maximize object coverage in cluttered scenes.
We keep only predictions whose confidence scores meet or exceed the threshold $\tau_{\text{OD-min-conf}}$.
The weighted boxes fusion (WBF) heuristic~\cite{DBLP:journals/ivc/SolovyevWG21} is used to merge multiple bounding boxes of the \textit{same} class into a single, confidence-weighted average box; boxes are considered overlapping if their intersection over union (IoU) exceeds a specified threshold $\tau_{\text{overlap-IoU}}$.\\
\noindent\textbf{\textit{Object segmentation}}
We subsequently refine object representations from rectangular bounding boxes to precise region masks with contour outlines.
When the segmenter fails, we fall back to using the box regions.

\paragraph{Relation Estimation.}
Having obtained object regions, we compute pairwise spatial relations between them--the \gom heart.
We create an ontology organizing 7~relation types into 3 groups: (i) directional (\texttt{above}, \texttt{below}, \texttt{left\_of}, \texttt{right\_of}), (ii) depth stacking (\texttt{in\_front\_of}, \texttt{behind}), and (iii) general proximity (\texttt{near}) for cases where directional classification is ambiguous.
To extend expressivity, (i)-type relations can be augmented with modifiers (\texttt{touching}, \texttt{very\_close}, \texttt{close}).

\noindent\textbf{\textit{Directional}}
For each object pair, we take the bounding box centers $(cx_i, cy_i)$ and $(cx_j, cy_j)$, and calculate the displacement vector  $(dx, dy) = (cx_j - cx_i, cy_j - cy_i)$.
Relations are determined by the dominant displacement direction: if $|dy| \geq |dx|$ and $|dy| > \tau_{\text{dir-margin}}$, the relation is \texttt{above} (if $dy < 0$) or \texttt{below} (if $dy > 0$).
If $|dx| > \tau_{\text{dir-margin}}$, the relation is \texttt{left\_of} or \texttt{right\_of} based on the sign of $dx$.

\noindent\textbf{\textit{Depth stacking}}
A monocular model recovers metric depth from $I$.
For each object pair, we sample the depth value at the bounding box centers, normalized to $[0,1]$ where higher values indicate objects nearer to the camera.
Depth relationships are established for all object pairs, even non-overlapped ones, with substantial z-difference: when $|\text{depth}_j - \text{depth}_i| > \tau_{\text{z-diff}}$, if $\text{depth}_j > \text{depth}_i$, object $j$ is \texttt{in\_front\_of} object $i$, otherwise \texttt{behind}.

\noindent\textbf{\textit{Proximity}}
If the center-based distance between two objects is below $\tau_{\text{near}}$ and they have not been assigned to a directional case, the \texttt{near} relation acts as fallback.

\noindent\textbf{\textit{Modifiers}}
We detail directional relations between objects of \textit{different} classes based on their closeness.
Objects are classified as \texttt{touching} when they exhibit overlap ($\text{IoU} > \tau_{\text{touch-IoU}}$) or minimal separation between box edges ($\leq \tau_{\text{touch-gap}}$).
For non-touching objects, we employ normalized distance $d_{\text{norm}}$--the Euclidean distance between centers scaled by $I$ dimensions--to establish proximity gradations: \texttt{very\_close} ($d_{\text{norm}} <~\tau_{\text{v-close}}$), \texttt{close} ($d_{\text{norm}} < \tau_{\text{close}}$).

\paragraph{Filtering}
The annotations computed so far are long-tailed and might incorporate information unrelated to the given task prompt $T$.
We thus deploy a two-step filtering pipeline.

\noindent\textbf{\textit{Objects}}
We retain only objects relevant to the query--those explicitly or implicitly mentioned through their labels, aliases, or synonyms.
For efficiency, the mention detection pipeline is incremental, progressing from lexical matching to semantic embedding comparison with cosine similarity $> \tau_{\text{query-obj}}$.
If multiple objects are relevant to the query, we retain only them and their interrelations.
If a single object is relevant, we keep that object, the relations for which it is the head, and the tail objects with which it interacts.
If no object is relevant, we retain all objects and relations to ensure comprehensive coverage when query specificity is insufficient.

\noindent\textbf{\textit{Relations}}
For each object, we retain only the top-$k$ relations having it as head.
Per-object ranking (the lower the better) employs a two-tier sorting, with query relevancy as the primary key (0 for relevant, 1 otherwise) and spatial distance as the secondary key.
In this way, for instance, depth relations between distant objects are naturally excluded in favor of nearby interactions.
A relation is judged relevant if its label matches at least one query relation term through the same matching pipeline introduced for the object case.
Query relation terms include mentioned paraphrases and synonyms of target relations.
To avoid redundancy, we iterate through the relation list (same order as presented) and retain only the first occurrence linking any two objects, thus preventing the modeling of both direct and inverse relations.

\paragraph{Scene Graph Rendering.}
Having defined meaningful objects (nodes) and spatial relations (edges), we possess all the ingredients to construct SGs overlaid on $I$.

\noindent\textbf{\textit{Node mark type}}
Object regions are rendered as mask marks with class-specific coloring (solid border, semi-transparent fill), ensuring objects of the same class share identical colors.
To enable MLM reference to specific objects in textual outputs, we render unique ID marks adjacent to each object with interpretable and speakable content.
We experiment with both numeric IDs and their textual extensions (formatted as \texttt{\textit{<class>}\_\textit{<id>}}).
ID marks appear within rectangular boxes sharing the border color of their corresponding object, with box fill and font color dynamically optimized for background contrast.

\noindent\textbf{\textit{Edge mark type}}
Relations are visualized as directed head-to-tail arrows, colored to match the head object.
In addition to connectivity, we investigate the impact of relation labels rendered as textual marks within boxes having the same border color as the arrow, white fill, and black font.

\noindent\textbf{\textit{Mark allocation}}
Effective mark (mask, ID/label, arrow) placement requires collision-free positioning to prevent MLM confusion.
The anti-overlapping strategy from SoM works exclusively to object IDs constrained within object regions, limiting its applicability to marks with minimal spatial footprint.
Since our approach incorporates textual object IDs and edge marks extending beyond single object regions, we propose a novel allocation algorithm.
First, we draw edge labels at object midpoints, as these marks are more intrusive visually.
Object ID marks are subsequently positioned at region centroids, remaining within boundaries when spatial constraints permit; otherwise, they are relocated externally while avoiding conflicts with existing ID and label marks.
Arrows are created with slightly shortened endpoints to accommodate textual labels.
When multiple arrows originate from the same object in similar directions, we increase their curvature radius progressively to prevent overlap.
Following initial placement, we use a resolver to iteratively displace conflicting marks along coordinate axes with small incremental steps until all collisions are eliminated.
To preserve visual coherence between displaced edge labels and their corresponding arrows, we render dashed lines connecting each label to its original midpoint position.

\subsection{Prompting}

We inspect two prompting modes, depending on the modalities involved in communicating spatial information.

\noindent\textbf{\textit{Visual SG}}
We populate $T$ with plain task instructions.
The MLM must interpret SG meaning solely through $I^{\text{SG}}$.

\noindent\textbf{\textit{Visual and Textual SG}}
We populate $T$ with both task instructions and a verbalized SG representation ($T^{\text{SG}}$), complementing $I^{\text{SG}}$.
We use triplet arrow notation\footnote{\texttt{\textit{<head>}--(\textit{<relation>})-->\textit{<tail>}}.} with interpretative guidance to facilitate MLM understanding.
The prompt template is reported in the Supplementary Material.

\noindent Following SoM, we emphasize that \gom supports tasks requiring visual outputs by enabling reverse mapping from spoken mark IDs back to their corresponding object regions.


\begin{figure*}[!htb]
\centering

\textbf{Question:} \textit{``Is the potted plant below the oven?''}

\vspace{0.2cm}
\begin{subfigure}[t]{0.32\textwidth}
    \centering
    \fontsize{8}{11}\selectfont
    \includegraphics[width=\textwidth]{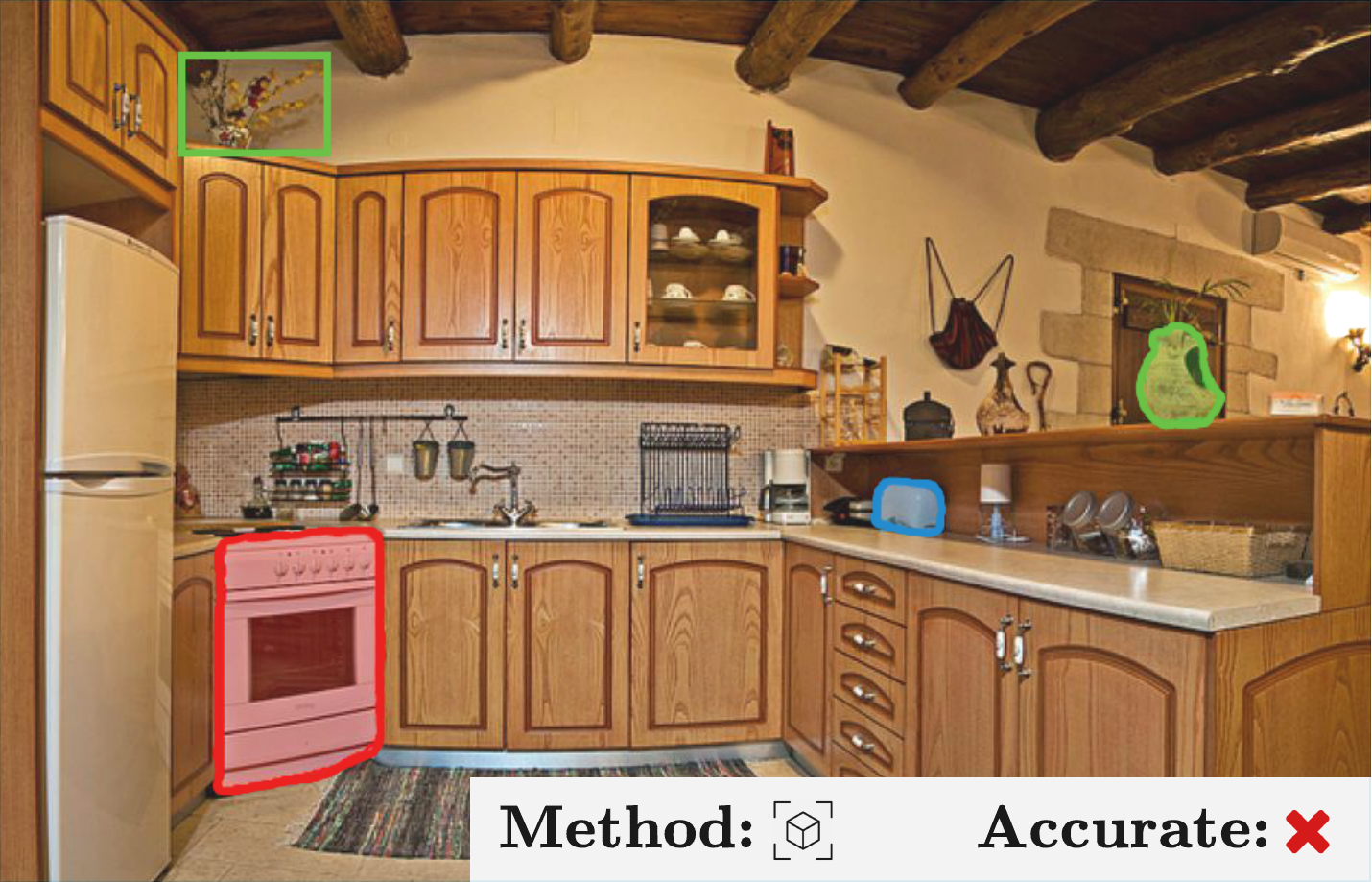}\\[-8pt]
    \colorbox{baseline}{%
    \begin{minipage}[t][1.8cm][t]{.95\textwidth}
        \vspace{-1pt}
        {\arrayrulecolor{tablebordercolor} 
        \renewcommand{\arraystretch}{1.5} 
        \begin{tabularx}{\textwidth}{X} 
        \textcolor{error}{Yes}, the potted plant is \textcolor{error}{below the oven}. It is located \textcolor{error}{on top of the refrigerator}. \\
        \end{tabularx}}
        \end{minipage}
    }
\end{subfigure}
\hfill
\begin{subfigure}[t]{0.32\textwidth}
    \centering
    \fontsize{8}{11}\selectfont
    \includegraphics[width=\textwidth]{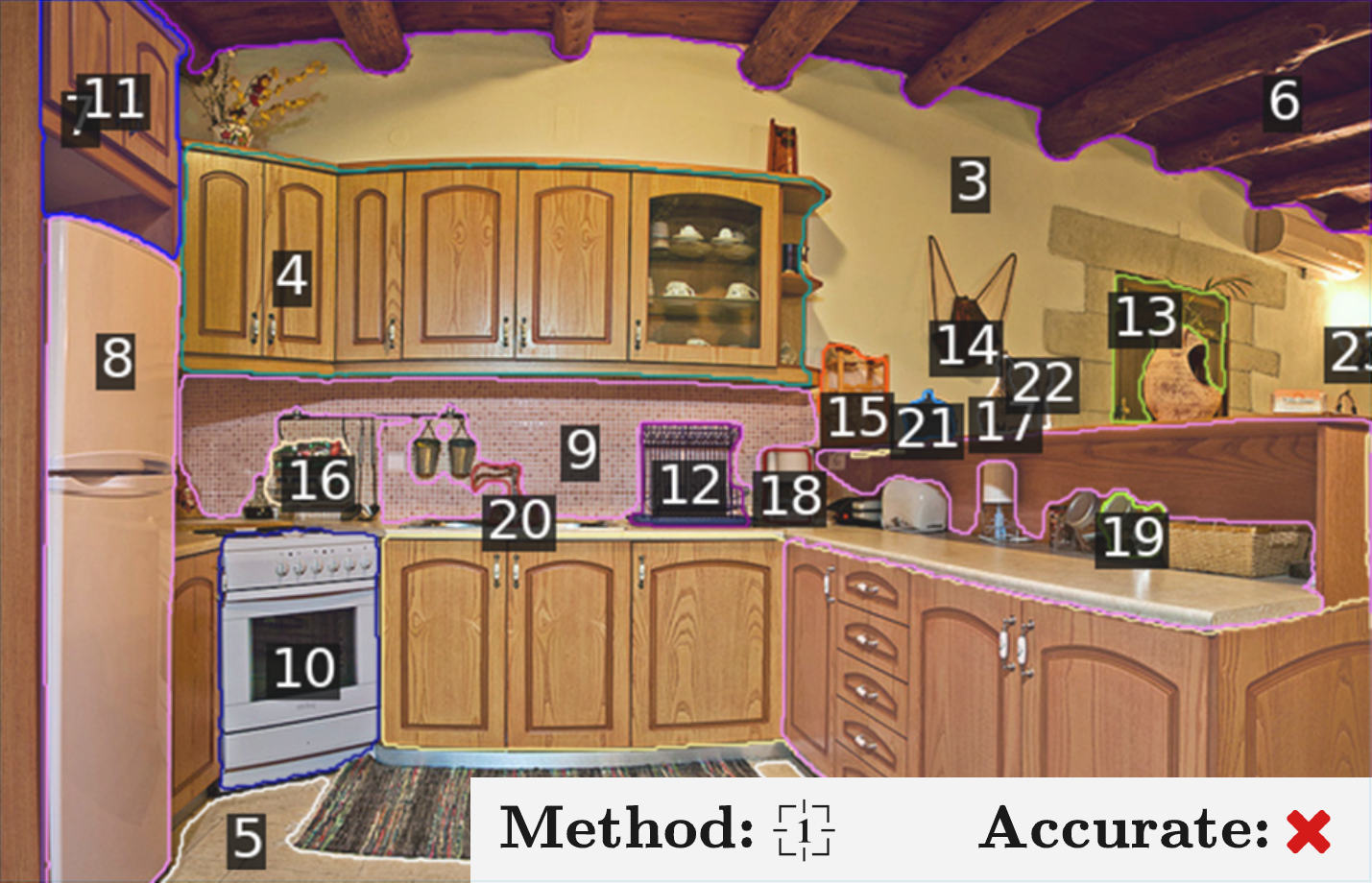}\\[-8pt]
    \colorbox{baseline}{%
    \begin{minipage}[t][1.8cm][t]{.95\textwidth}
        \vspace{-1pt}
        {\arrayrulecolor{tablebordercolor} 
        \renewcommand{\arraystretch}{1.5} 
        \begin{tabularx}{\textwidth}{X} 
        \textcolor{error}{Yes}, the potted plant is \textcolor{error}{below the oven}. According to the image, the potted plant (\textcolor{error}{labeled as number 16}) is positioned \textcolor{error}{below and to the left of the oven}. \\
        \end{tabularx}}
        \end{minipage}
    }
\end{subfigure}
\hfill
\begin{subfigure}[t]{0.32\textwidth}
    \centering
    \fontsize{8}{11}\selectfont
    \includegraphics[width=\textwidth]{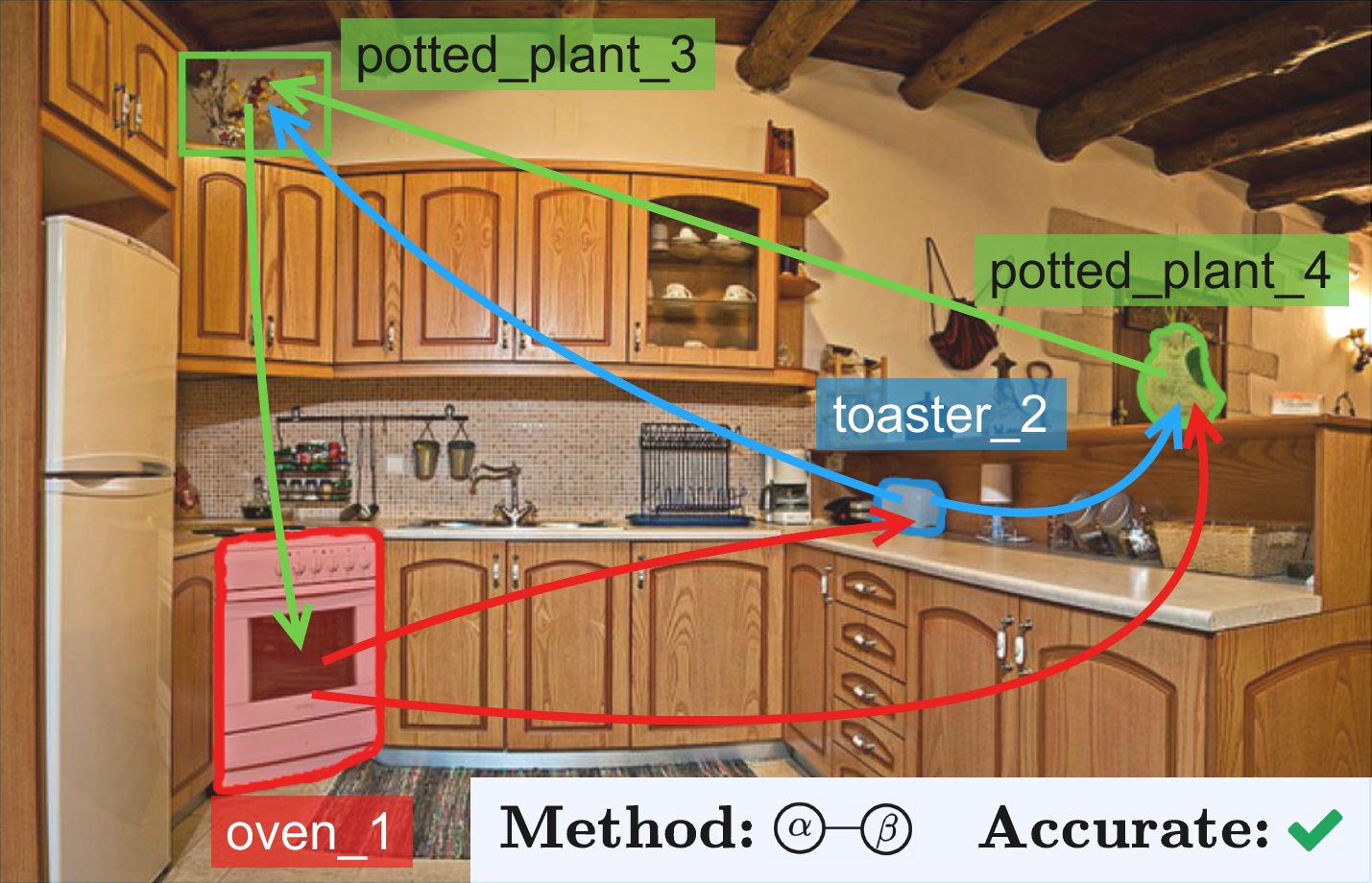}\\[-8pt]
    \colorbox{gom}{%
    \begin{minipage}[t][1.8cm][t]{.95\textwidth}
        \vspace{-1pt}
        {\arrayrulecolor{tablebordercolor} 
        \renewcommand{\arraystretch}{1.5} 
        \begin{tabularx}{\textwidth}{X} 
        No, the potted plant is not below the oven. The potted plant is located on the countertop to the right of the oven, and it is above the oven in the image. \\
        \end{tabularx}}
        \end{minipage}
    }  
\end{subfigure}

\vspace{0.2cm}

\begin{subfigure}[t]{0.32\textwidth}
    \centering
    \fontsize{8}{11}\selectfont
    \includegraphics[width=\textwidth]{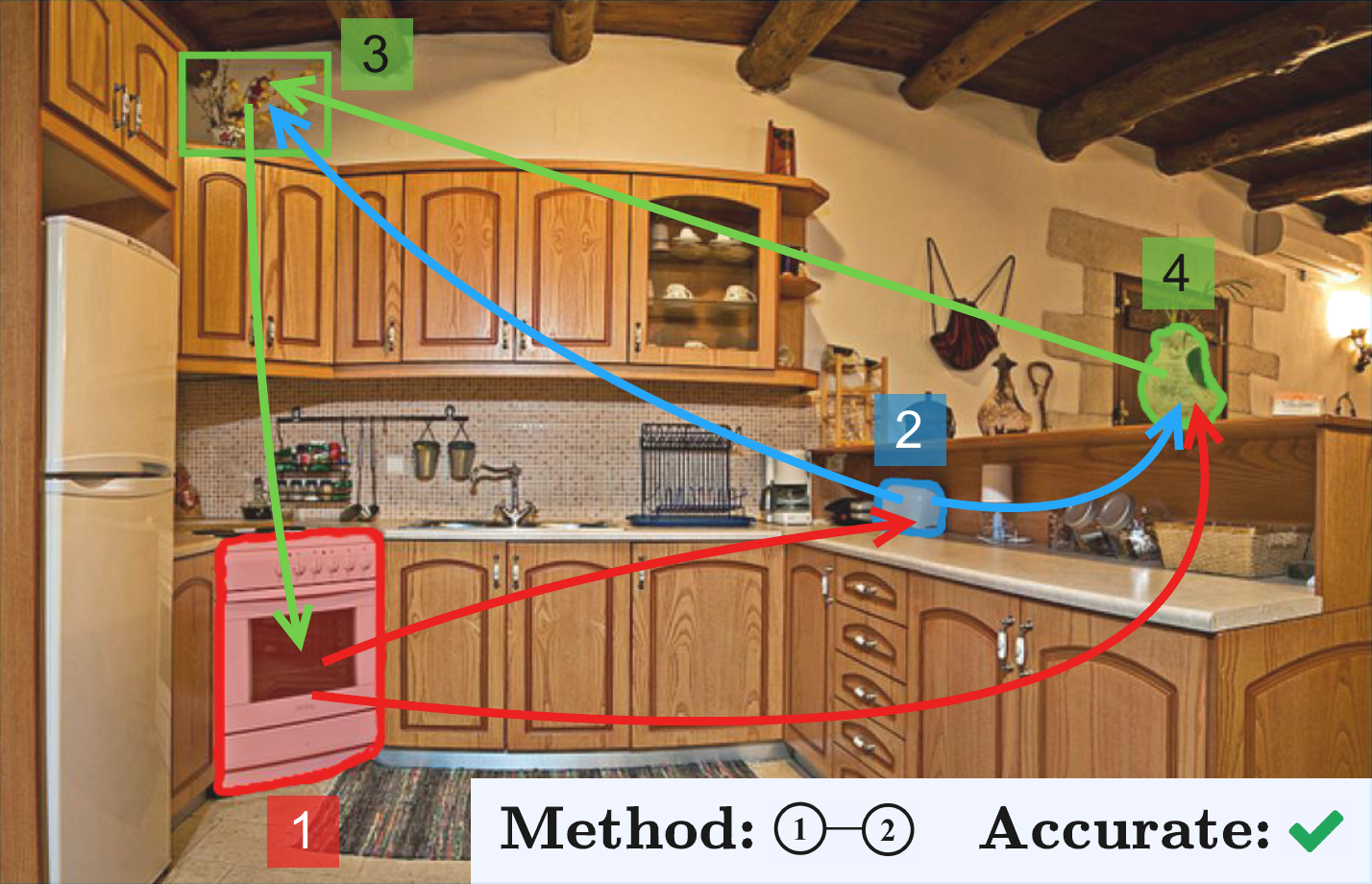}\\[-8pt]
    \colorbox{gom}{%
    \begin{minipage}[t][2.1cm][t]{.95\textwidth}
        \vspace{-1pt}
        {\arrayrulecolor{tablebordercolor} 
        \renewcommand{\arraystretch}{1.5} 
        \begin{tabularx}{\textwidth}{X} 
        No. According to the image, the potted plant (labeled as number 4) is located on a countertop to the right of the oven, and it is positioned above the oven when considering the vertical alignment. \\
        \end{tabularx}}
        \end{minipage}
    }
\end{subfigure}
\hfill
\begin{subfigure}[t]{0.32\textwidth}
    \centering
    \fontsize{8}{11}\selectfont
    \includegraphics[width=\textwidth]{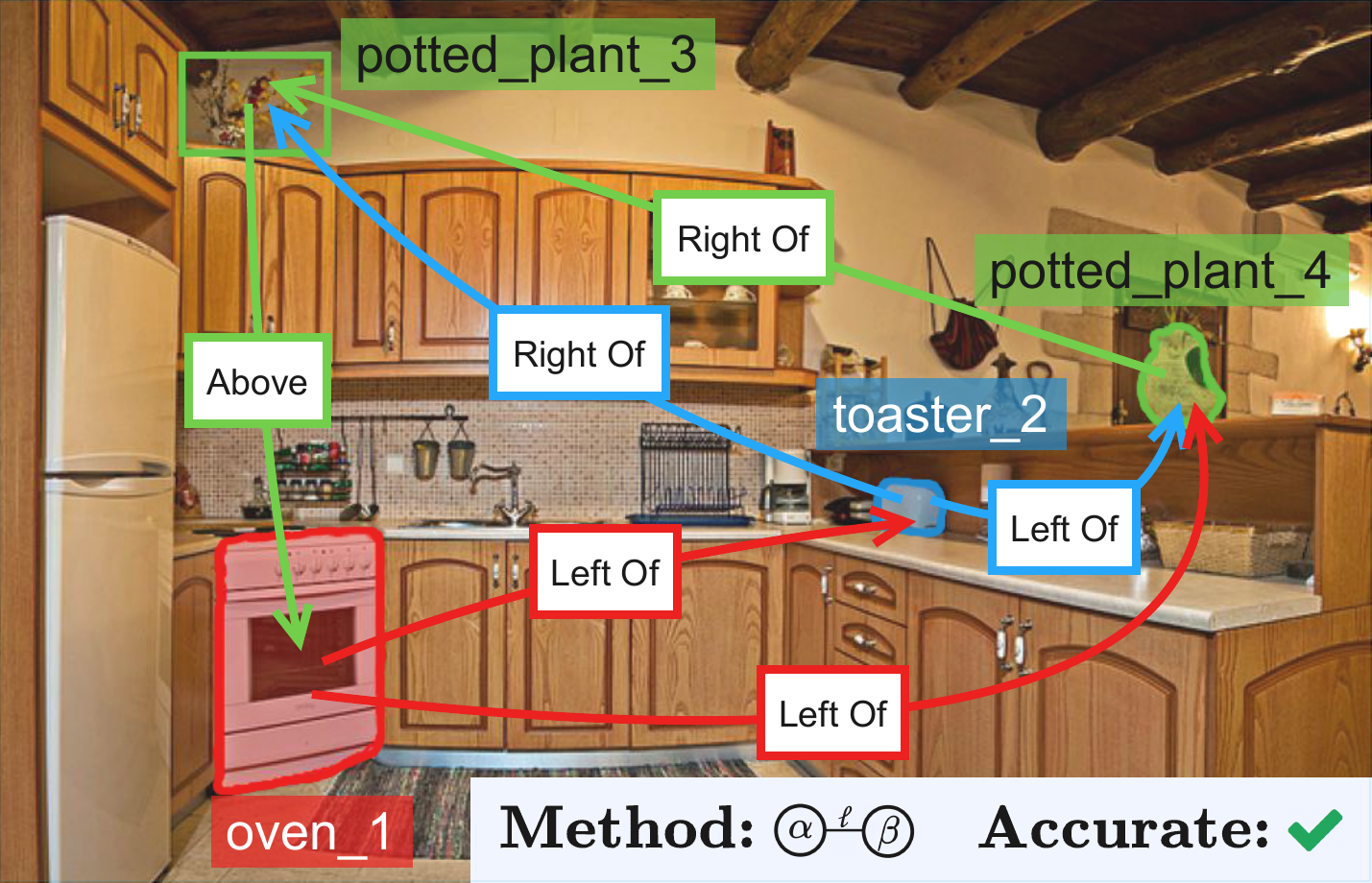}\\[-8pt]
    \colorbox{gom}{%
    \begin{minipage}[t][2.1cm][t]{.95\textwidth}
        \vspace{-1pt}
        {\arrayrulecolor{tablebordercolor} 
        \renewcommand{\arraystretch}{1.5} 
        \begin{tabularx}{\textwidth}{X} 
        No, the potted plant is not below the oven. According to the image, the potted plant (labeled as \quotes{plotted plant\_3}) is positioned above the oven (labeled as \quotes{oven\_1}). \\
        \end{tabularx}}
        \end{minipage}
    }
\end{subfigure}
\hfill
\begin{subfigure}[t]{0.32\textwidth}
    \centering
    \fontsize{8}{11}\selectfont
    \includegraphics[width=\textwidth]{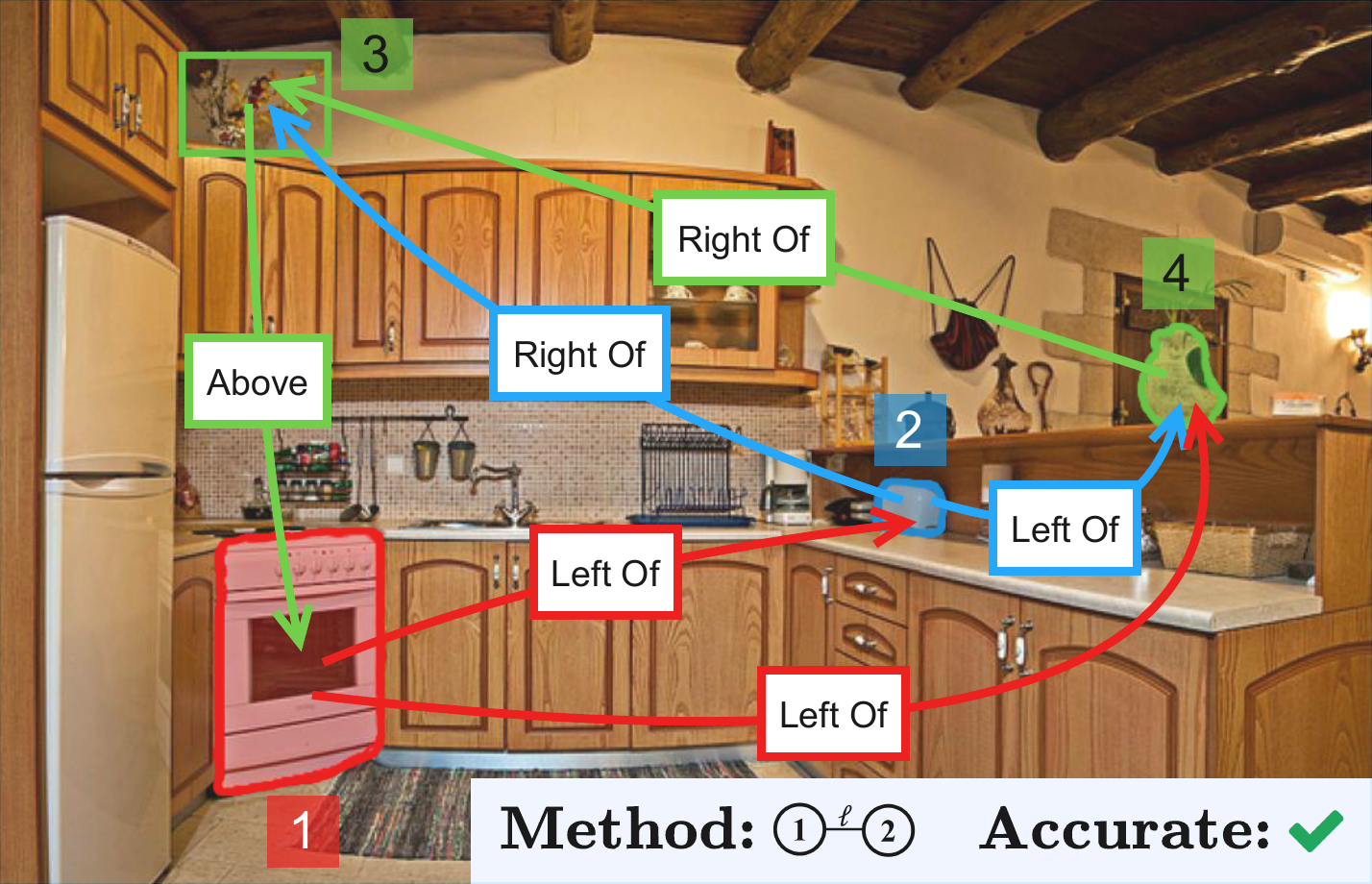}\\[-8pt]
    \colorbox{gom}{%
    \begin{minipage}[t][2.1cm][t]{.95\textwidth}
        \vspace{-1pt}
        {\arrayrulecolor{tablebordercolor} 
        \renewcommand{\arraystretch}{1.5} 
        \begin{tabularx}{\textwidth}{X} 
        No, the potted plant is not below the oven. According to the image, the potted plant (labeled as 4) is positioned to the right of the oven and above it. \\
        \end{tabularx}}
        \end{minipage}
    }
\end{subfigure}

\caption{\textbf{Qualitative example illustrating the impact of image preprocessing on VQA performance.} The same question from VQA-v2 is posed to Qwen2.5-7B using 6 different hard visual prompts, highlighting how pixel transformations can influence the model's responses. For figure readability, the font size and line thickness have been increased compared to their actual values. Gray boxes denote baseline outputs, while blue boxes indicate those from our proposed \gom. See icon legend in Table~\ref{tab:main_results}.}
\label{fig:vqa_preprocessing}
\end{figure*}


\section{Experimental Setup}

\subsection{Datasets and Metrics}

We evaluate \gom's performance across a suite of VQA and REC tasks reflecting complementary dimensions of spatial visual understanding.
For VQA, we include GQA~\cite{DBLP:conf/cvpr/HudsonM19}, VQAv1~\cite{DBLP:conf/iccv/AntolALMBZP15}, and VQAv2~\cite{DBLP:conf/cvpr/GoyalKSBP17}.
All three feature open-ended questions and are evaluated using accuracy-based metrics following the official evaluation protocols defined by their respective creators.
GQA demands multi-step spatial reasoning and fine-grained object relationships.
VQAv1 covers general queries that combine vision, language, and commonsense reasoning.
VQAv2 mitigates language bias by pairing visually similar images with different answers to identical questions.
For REC, we analyze RefCOCOg~\cite{DBLP:conf/emnlp/KazemzadehOMB14}, which involves localizing a specific object instance based on a textual description, often including spatial details to disambiguate the referent.
We approach the task by prompting the MLM to generate the ID of the target object; a prediction is considered correct if the region associated with the predicted object has an $\text{IoU} \geq 0.9$ with the ground truth bounding box.
We underline that this evaluation setup is not applicable when the model receives only $I$ or its segmentation-augmented version.
In line with established practices~\cite{DBLP:journals/corr/abs-2310-11441}, we randomly sample 1K images from each dataset to manage computational and time constraints. 
In contrast to prior work that filters specific queries, we retain all questions associated with each selected image, ensuring robust and statistically meaningful evaluation.
On average, each image in GQA, VQAv1, VQAv2, and RefCOCOg is paired with 3, 4, 4, and 1 queries, respectively.

\subsection{Implementation Details}

\paragraph{Object Detectors.}
Modern object detection has evolved to support both closed-vocabulary (CV) and open-vocabulary (OV) scenarios, with models increasingly expected to handle unknown classes and generalize across domains~\cite{DBLP:journals/tcsv/LiWWLLY25}. 
While OV detectors offer strong flexibility through text-conditioned recognition and self-supervised training, they often trade off precision or require specialized prompt design.
To maximize object recall in our pipeline without additional training, we combine the outputs of three complementary, training-free detectors. 
These include OWL-V2~\cite{li2024owlv2}, an OV model capable of detecting arbitrary categories via textual prompts; YOLOv8-X~\cite{ultralytics2023yolov8}, a fast and high-confidence CV detector for common classes; and Mask R-CNN R101-FPN from Detectron2~\cite{wu2019detectron2}, CV, which supplements robust region proposals, especially for persons.

\begin{table}[!tb]
\centering
\fontsize{9}{11}\selectfont
\begin{threeparttable}
\begin{tabularx}{\columnwidth}{@{}p{1cm}X@{}}
\toprule
\addlinespace[-0.1pt]
\textbf{Phase} & \textbf{Hyperparam Setting} \\[-0.5mm]
\midrule
\faObjectUngroup & $\tau_{\text{OD-min-conf}}=\{0.4, 0.5\ast, 0.8\}\; \tau_{\text{overlap-IoU}}=0.9$ \\
{\large\ding{182}/\ding{183}/\ding{184}} & \makecell[l]{$\tau_{\text{dir-margin}\tnote{$\ddagger$}}=\{10, 20\ast, 50\},\; \tau_{\text{z-diff}}=\{0.1\ast, 0.15, 0.20\},$\\ $\tau_{\text{near}\tnote{$\ddagger$}}=5,000$} \\
\faTag & \makecell[l]{$\tau_{\text{touch-IoU}\tnote{$\ddagger$}}=0.1,\; \tau_{\text{touch-gap}\tnote{$\ddagger$}}=3,\;$\\$\tau_{\text{v-close}\tnote{$\ddagger$}}=0.05,\; \tau_{\text{close}\tnote{$\ddagger$}}=\{0.12\ast, 0.15\}$} \\
\faCube/\faArrowsH & $\tau_{\text{query-obj}}=\{0.5\ast, 0.7, 0.8\},\; k=\{1, 2, 3\ast, 4, 5, 6\}$ \\
\cdashline{1-2}
\faEye & \makecell[l]{$\text{seed}=\{42, 123, 456\},$\\$\text{temp}=\{0.1, 0.3, 0.5\},\; \text{top-p}=\{0.7, 0.9, 0.95\}$}\\
\bottomrule
\end{tabularx}
\begin{tablenotes}
\item[$\ddagger$] Normalized pixel-level distance.
\item[] \textbf{\textit{Symbols}:} \faObjectUngroup\,= Object detection; {\large\ding{182}}/{\large\ding{183}}/{\large\ding{184}}\,= Directional/Depth/Proximity; \faTag\,= Modifiers; \faCube/\faArrowsH\,= Object/Relation filtering; \faEye\,= Decoding.
\end{tablenotes}
\vspace{-0.1cm}
\end{threeparttable}
\caption{\textbf{Hyperparameter sweep.} Top: algorithm. Bottom: decoding (for best algorithm config). $\ast$ = algorithm values, picked after preliminary runs with Qwen-2.5-VL on GQA.\vspace{-0.4cm}}
\label{tab:hyperparameters}
\end{table}

\newpage
\paragraph{Segmenter.}  
Transformer-based segmentation models have become the standard for dense prediction tasks, offering strong performance through unified, query-driven architectures~\cite{DBLP:journals/pami/LiDYZPCCLL24}.
We adopt Segment Anything SAM-HQ~\cite{ke2023samhq}, OV, which produces high-resolution instance masks without retraining.
Its promptable design and use-case robustness make it suitable for isolating fine object boundaries from unconstrained image sources.

\paragraph{Depth Estimator.}  
Monocular depth estimation has seen rapid progress, with recent models targeting metric depth recovery in scale- and shift-invariant settings~\cite{DBLP:conf/cvpr/YangKHXFZ24,DBLP:conf/cvpr/WangXDX00Y25}. 
However, these approaches often rely on accurate focal length and calibration parameters, which are hard to obtain for unconstrained, in-the-wild images and add significant preprocessing overhead.
For this reason, we employ MiDaS DPT-Large~\cite{ranftl2022midas}, which estimates relative depth without requiring camera metadata.
Trained on a broad range of indoor and outdoor scenes, MiDaS provides reliable predictions in open-world settings while also offering low memory usage and latency.

\paragraph{Filtering.}
Aliases and synonyms for object labels are derived through WordNet (v3.0) synsets.
The semantic recognition of object mentions within the query draws on pretrained FastText embeddings (cc.en.300.vec).

\paragraph{MLMs.}  
We test Qwen-2.5-VL-7B (Instruct)~\cite{DBLP:journals/corr/abs-2502-13923}, Gemma-3-4B (Instruct)~\cite{DBLP:journals/corr/abs-2503-19786}, and LlamaV-o1-11B~\cite{DBLP:journals/corr/abs-2501-06186}.
These open-source models were selected to ensure empirical diversity between architectures, sizes, and usage scenarios, with a focus on common low- and mid-resource settings.
Our aim is to verify whether \gom improves downstream performance without relying on instruction-heavy or domain-specialized models.
LlamaV-o1, a reasoning model, allows us to inquire whether \gom adds value to step-by-step thoughts.

\paragraph{Hyperparameters.}
Table~\ref{tab:hyperparameters} lists the hyperparameters along with their empirical search grid.
After selecting the optimal algorithmic configuration, we run MLM inference using nucleus sampling and capping generation at 512 tokens.
Each run ($<$MLM, prompt method, dataset$>$) is repeated using 27 combinations of seed, temperature, and top-p.

\paragraph{Hardware Setup.}
All runs were conducted on a workstation running Ubuntu 20.04.3 LTS, equipped with a single NVIDIA GeForce RTX3090 GPU (24GB VRAM), 64GB of RAM, and an Intel® Core™ i9-10900X CPU @ 3.70GHz.

\begin{table}[!tb]
\centering
\fontsize{9}{11}\selectfont
\begin{tabular*}{\columnwidth}{@{\extracolsep{\fill}}llcccc}
\toprule
\addlinespace[-0.1pt]
\fontsize{9}{11}\selectfont
\faEye & \textbf{Method} & \textbf{GQA} & \multicolumn{2}{c}{\textbf{VQA}} & \textbf{RefCOCOg} \\[-1mm]
\cmidrule(lr){4-5}
\addlinespace[-0.3pt]
& & & \textbf{v1} & \textbf{v2} & \\[-1mm]
\midrule
\multirow{7}{*}{\raisebox{-65pt}{\rotatebox{90}{\textbf{Gemma-3\ 4B}}}}
& \rawImage[0.15in] & 56.2$_{\pm 0.11}$ & 64.3$_{\pm 0.14}$ & 59.9$_{\pm 0.19}$ & -- \\
& \seg[0.15in] & 53.8$_{\pm 0.34}$ & 63.7$_{\pm 1.15}$ & 60.2$_{\pm 1.37}$ & -- \\
& \segNumObj[0.15in] & 56.9$_{\pm 0.34}$ & 63.8$_{\pm 0.07}$ & 59.0$_{\pm 0.18}$ & 54.8$_{\pm 0.91}$ \\[3pt]
\cdashline{2-6}\\[-2.5mm]
& \gomLabObj[0.4in] & 58.8$_{\pm 0.16}$ & 65.2$_{\pm 0.38}$ & 62.8$_{\pm 0.36}$ & 56.3$_{\pm 0.91}$ \\
& \gomNumObj[0.4in] & 60.3$_{\pm 0.45}$ & 71.5$_{\pm 0.1}$ & 70.2$_{\pm 0.25}$ & \cellcolor{cellhighlight}\textbf{56.4$_{\pm 0.66}$} \\
& \gomLabObjLabRel[0.4in] & \cellcolor{cellhighlight}\textbf{63.2$_{\pm 0.07}$} & \cellcolor{cellhighlight}\textbf{74.2$_{\pm 0.16}$} & \cellcolor{cellhighlight}\textbf{71.9$_{\pm 0.01}$} & 56.3$_{\pm 0.58}$ \\
& \gomNumObjLabRel[0.4in] & 61.2$_{\pm 0.28}$ & 71.2$_{\pm 0.07}$ & 70.2$_{\pm 0.15}$ & 56.3$_{\pm 0.42}$ \\[2pt]
\midrule
\multirow{7}{*}{\raisebox{-75pt}{\rotatebox{90}{\textbf{Qwen-2.5-VL\ 7B}}}}
& \rawImage[0.15in] & 61.6$_{\pm 0.11}$ & 77.7$_{\pm 0.11}$ & 73.8$_{\pm 0.16}$ & -- \\
& \seg[0.15in] & 53.2$_{\pm 0.30}$ & 65.2$_{\pm 1.21}$ & 67.2$_{\pm 1.20}$ & -- \\
& \segNumObj[0.15in] & 61.8$_{\pm 0.35}$ & 65.4$_{\pm 0.3}$ & 68.6$_{\pm 0.16}$ & 55.5$_{\pm 0.39}$ \\[3pt]
\cdashline{2-6}\\[-2.5mm]
& \gomLabObj[0.4in] & 62.5$_{\pm 0.32}$ & 77.9$_{\pm 0.42}$ & 74.0$_{\pm 0.06}$ & 56.5$_{\pm 0.90}$ \\
& \gomNumObj[0.4in] & 61.9$_{\pm 0.41}$ & 78.1$_{\pm 0.37}$ & 76.4$_{\pm 0.21}$ & 56.8$_{\pm 0.74}$ \\
& \gomLabObjLabRel[0.4in] & 63.8$_{\pm 0.13}$ & \cellcolor{cellhighlight}\textbf{82.5$_{\pm 0.48}$} & \cellcolor{cellhighlight}\textbf{80.5$_{\pm 0.03}$} & 57.4$_{\pm 0.21}$ \\
& \gomNumObjLabRel[0.4in] & \cellcolor{cellhighlight}\textbf{65.0$_{\pm 0.1}$} & 81.1$_{\pm 0.1}$ & 80.1$_{\pm 0.43}$ & \cellcolor{cellhighlight}\textbf{57.5$_{\pm 0.65}$} \\[2pt]
\midrule
\multirow{7}{*}{\raisebox{-70pt}{\rotatebox{90}{\textbf{LlamaV-o1\ 11B}}}}
& \rawImage[0.15in] & 60.2$_{\pm 0.20}$ & 75.3$_{\pm 0.11}$ & 75.1$_{\pm 0.12}$ & -- \\
& \seg[0.15in] & 61.9$_{\pm 0.35}$ & 72.6$_{\pm 1.22}$ & 75.3$_{\pm 1.22}$ & -- \\
& \segNumObj[0.15in] & 62.0$_{\pm 0.33}$ & 72.8$_{\pm 0.42}$ & 75.5$_{\pm 0.38}$ & 55.3$_{\pm 1.03}$ \\[3pt]
\cdashline{2-6}\\[-2.5mm]
& \gomLabObj[0.4in] & 62.4$_{\pm 0.3}$ & 76.0$_{\pm 0.49}$ & 75.9$_{\pm 0.23}$ & 57.2$_{\pm 0.79}$ \\
& \gomNumObj[0.4in] & 62.5$_{\pm 0.3}$ & 77.9$_{\pm 0.42}$ & 80.9$_{\pm 0.33}$ & 57.3$_{\pm 0.71}$ \\
& \gomLabObjLabRel[0.4in] & \cellcolor{cellhighlight}\textbf{67.0$_{\pm 0.44}$} & 79.8$_{\pm 0.31}$ & 83.4$_{\pm 0.02}$ & 56.9$_{\pm 0.94}$ \\
& \gomNumObjLabRel[0.4in] & 65.0$_{\pm 0.37}$ & \cellcolor{cellhighlight}\textbf{83.2$_{\pm 0.08}$} & \cellcolor{cellhighlight}\textbf{83.6$_{\pm 0.04}$} & \cellcolor{cellhighlight}\textbf{57.6$_{\pm 0.26}$} \\[2pt]
\bottomrule
\end{tabular*}\\[1mm]
{\arrayrulecolor{tablebordercolor}
\begin{tabularx}{\columnwidth}{|p{1.5cm}X|}
\hline
\rowcolor{tablehighlight} \multicolumn{2}{|l|}{\textcolor{tablebordercolor}{\textbf{Prompt Legend}}} \\
\rowcolor{tablehighlight} \textit{Baselines} & \rawImage[0.15in] Raw image, \seg[0.15in] Segmented objects, \segNumObj[0.15in] Segmented objects + Object Num IDs (SoM) \\
\hline
\rowcolor{tablehighlight} \gom \textit{(Ours)} & \makecell[l]{Segmented objects + Relations +\\ $\text{\gomLabObj[0.4in] Object Text IDs}\;/$\\
$\text{\gomNumObj[0.4in] Object Num IDs}\;/$\\
$\text{\gomLabObjLabRel[0.4in] Object Text IDs + Relation labels}\;/$\\
$\text{\gomNumObjLabRel[0.4in] Object Num IDs + Relation labels}$} \\
\hline
\end{tabularx}}
\caption{\textbf{Accuracy results ($\uparrow$).} Comparison between \gom variants (Visual SG only) and baseline prompt strategies on 4 datasets and 3 MLMs. We report mean $\pm$ standard deviation across decoding runs. Best results are highlighted.}
\label{tab:main_results}
\end{table}

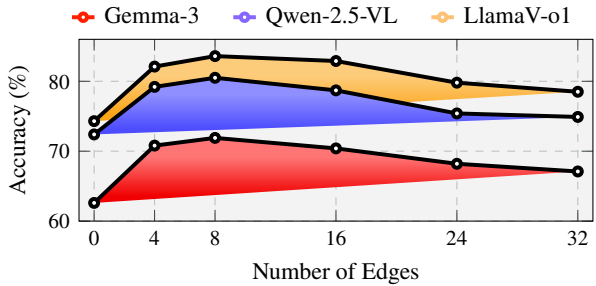
\begin{figure}[!htb]
    \centering
    \begin{subfigure}[t]{\linewidth}
        \begin{tikzpicture}
            \footnotesize
            \node[draw=none, inner sep=2pt, align=center, text width=\linewidth] {
            \begin{tabular}{lll}
                \tikz{\draw[color=red!75!orange, line width=1.5pt] (0,0) -- (0.3,0) node[pos=0.5, circle, fill=white, draw=red!75!orange, inner sep=1pt] {};} Gemma-3 & 
                \tikz{\draw[color=blue!80!magenta!60!white, line width=1.5pt] (0,0) -- (0.3,0) node[pos=0.5, circle, fill=white, draw=blue!80!magenta!60!white, inner sep=1pt] {};} Qwen-2.5-VL &
                \tikz{\draw[color=yellow!10!orange!60, line width=1.5pt] (0,0) -- (0.3,0) node[pos=0.5, circle, fill=white, draw=yellow!10!orange!60, inner sep=1pt] {};} LlamaV-o1
            \end{tabular}
            };
        \end{tikzpicture}
    \end{subfigure}
    \begin{subfigure}[H]{\linewidth}
        \begin{tikzpicture}
        \begin{axis}[
            width=\linewidth, height=4cm,
            ymajorgrids=true,
            xmajorgrids=true,
            grid=both,
            grid style=dashed,
            axis background/.style={fill=plotbackground},
            xmin=-1, xmax=33,
            xtick={0,4,8,16,24,32},
            xlabel={Number of Edges},
            xlabel style={font=\small},
            ymin=60, ymax=86,
            ylabel={Accuracy (\%)},
            ylabel style={font=\small},
            every tick label/.append style={font=\fontsize{8}{8}\selectfont},
        ]
        
        \addplot [
            shading=redgradient,
            line width=1.5pt,
            mark=*,
            mark size=1.5pt,
            mark options={fill=white},
        ] coordinates {
            (0, 62.6)
            (4, 70.8)
            (8, 71.9)
            (16, 70.4)
            (24, 68.2)
            (32, 67.1)
        }; \label{plot:gemma_edges}
        
        \addplot [
            shading=orangegradient,
            line width=1.5pt,
            mark=*,
            mark size=1.5pt,
            mark options={fill=white},
        ] coordinates {
            (0, 74.3)
            (4, 82.1)
            (8, 83.6)
            (16, 82.9)
            (24, 79.8)
            (32, 78.5)
        }; \label{plot:llama_edges}
        
        \addplot [
            shading=purplegradient,
            line width=1.5pt,
            mark=*,
            mark size=1.5pt,
            mark options={fill=white},
        ] coordinates {
            (0, 72.4)
            (4, 79.2)
            (8, 80.5)
            (16, 78.7)
            (24, 75.4)
            (32, 74.9)
        }; \label{plot:qwen_edges}
        
        \end{axis}
        \end{tikzpicture}
    \end{subfigure}
    \caption{\textbf{Effect of graph density.} Performance of \gom in VQAv2 as a function of the number of edges in the visual scene graph. $0$ edges corresponds to SoM-like prompting.\vspace{-0.08cm}}
    \label{fig:edge_ablation}
\end{figure}

\section{Results}

We evaluate \gom along three axes: (1) impact on spatial reasoning performance, (2) benefits of multimodal SG integration, and (3) efficiency in $I^{\text{SG}}$ computation.
Fair comparisons are made against vanilla MLMs and competitive hard visual prompting techniques, i.e., segmentation-only and SoM.

\paragraph{Graph Guidance in Spatial Reasoning.}

Main results are shown in Table~\ref{tab:main_results}.
\gom demonstrates superior performance across all experimental conditions, substantiating the efficacy of explicit relational encoding for spatial reasoning and scene comprehension.
Zooming out, we elucidate that even lightweight open-source MLMs with $\leq$11B parameters can adeptly exploit hard visual prompting, contravening precedent publications that documented success solely within commercial closed-source models such as GPT-4V~\cite{DBLP:journals/corr/abs-2310-11441}.
Gemma-3 experiences the most pronounced improvement, whereas Qwen manifests adverse sensitivity to SoM, frequently failing to reference the correct object regions.
LlamaV attains the highest absolute scores--83.6 in VQA and 57.6 in REC--suggesting that reasoning models are particularly adept at leveraging \gom representations.
Notably, despite the absence of a universally optimal \gom variant, the basic setting featuring textual object IDs without relation labels generally achieves top performance in VQA, while numeric IDs are desirable for REC.
Targeted error analysis reveals that models track textual descriptors more faithfully when generating abstractive answers.
When relational labels are omitted, models systematically ignore directional indicators.
Figure~\ref{fig:vqa_preprocessing} presents qualitative input-output augmentation examples.
Only graph-enhanced images enable accurate spatial interpretation; SoM conversely degrades reasoning through erroneous regional attributions.
Maximal \gom effectiveness materializes with 3-10 entities and 4-16 relations (Figure~\ref{fig:edge_ablation}).
Beyond this range, surplus annotations introduce noise, diminishing the headroom.

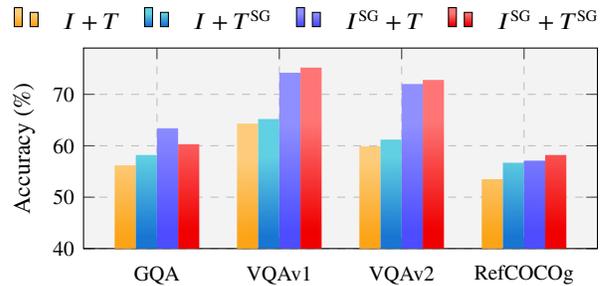
\begin{figure}[!htb]
    \centering
    \begin{subfigure}[t]{\linewidth}
        \hspace{-3mm}
        \begin{tikzpicture}
            \footnotesize
            \node[draw=none, align=center] {
                \setlength{\tabcolsep}{0.16cm}
                \begin{tabular}{llllllll}
                     \ref{plot:raw_image_raw_text} & $I + T$ & \ref{plot:raw_image_sg_text} & $I + T^{\text{SG}}$ & \ref{plot:sg_image_raw_text} & $I^\text{SG} + T$ & \ref{plot:sg_image_sg_text} &  $I^\text{SG}$ + $T^{\text{SG}}$
                \end{tabular}
            };
        \end{tikzpicture}
    \end{subfigure}
    \begin{subfigure}[H]{\linewidth}
        \begin{tikzpicture}
        \begin{axis}[
            width=1.00\linewidth, height=4.25cm,
            ymajorgrids=true,
            grid=both,
            grid style=dashed,
            axis background/.style={fill=plotbackground},
            ybar=0pt, 
            bar width=8pt,
            enlarge x limits=0.2,  
            xtick=data,
            symbolic x coords={
                GQA,
                VQAv1,
                VQAv2,
                RefCOCOg
            },
            xticklabel style={rotate=0, anchor=center, yshift=-10pt},
            xlabel style={font=\small},
            ymax=79,
            ymin=40,
            ylabel={Accuracy (\%)},
            ylabel style={font=\small},
            every tick label/.append style={font=\fontsize{8}{8}\selectfont},
            xtick pos=both,
            xtick align=inside,
        ]
        
        \addplot [shading=orangegradient, draw=none, error bars/.cd, y dir=both, y explicit] coordinates {
            (GQA, 56.2)
            (VQAv1, 64.3)
            (VQAv2, 59.9)
            (RefCOCOg, 53.5)
        };\label{plot:raw_image_raw_text}
        
        \addplot [shading=bluegradient, draw=none, error bars/.cd, y dir=both, y explicit] coordinates {
            (GQA, 58.2)
            (VQAv1, 65.2)
            (VQAv2, 61.2)
            (RefCOCOg, 56.7)
        };\label{plot:raw_image_sg_text}
        
        \addplot [shading=purplegradient, draw=none, error bars/.cd, y dir=both, y explicit] coordinates {
            (GQA, 63.4)
            (VQAv1, 74.2)
            (VQAv2, 72.0)
            (RefCOCOg, 57.1)
        };\label{plot:sg_image_raw_text}
        
        \addplot [shading=redgradient, draw=none, error bars/.cd, y dir=both, y explicit] coordinates {
            (GQA, 60.3)
            (VQAv1, 75.2)
            (VQAv2, 72.8)
            (RefCOCOg, 58.2)
        };\label{plot:sg_image_sg_text}
        
        \draw (axis cs:{[normalized]\pgfkeysvalueof{/pgfplots/xmin}},0)
            -- (axis cs:{[normalized]\pgfkeysvalueof{/pgfplots/xmax}},0);
        
        \end{axis}
        \end{tikzpicture}
    \end{subfigure}
    \caption{\textbf{Accuracy impact deriving from augmenting the visual ($I$) and textual ($T$) prompt with scene graphs (SG apex).} Gemma-3 results. Proposed \gom solutions have $I^\text{SG}$.\vspace{-0.01cm}}
    \label{fig:vision-desc-comparison}
\end{figure}

\vspace{0.7cm}
\paragraph{Scene Graph Modality.}
Figure~\ref{fig:vision-desc-comparison} systematically evaluates the contribution of SGs across distinct prompt modalities.
Visual graph representations repeatedly lead to higher accuracy compared to textual encodings exclusively examined in prior work, with elevations of up to +10\%. 
Remarkably, the combination of verbalized SGs with their visual counterparts always produces performance gains, albeit of modest magnitude.
This empirical evidence corroborates that \gom enhancements stem not from the superficial presence of graph structures--as postulated by other research groups--but rather from a deliberate visual graph prompting pipeline that better activates the latent reasoning faculties of MLMs.

\newpage
\paragraph{Efficiency.}
The performance-efficiency tradeoff is a critical consideration habitually neglected in evaluations of visual prompting methodologies.
Comparative latency analysis on our computational infrastructure reveals that \gom incurs very low overhead, averaging 1.13 seconds per image compared to 0.77 and 0.92 seconds for segmentation-only and SoM approaches, respectively, attributable to relation estimation.
On the other hand, this cost is offset by advances in spatial reasoning, primarily in VQA tasks.


\section{Conclusion}

We introduced Graph-of-Mark, the first visual prompting method that embeds depth-aware scene graphs into the input image to support spatial reasoning. 
Unlike prior techniques that treat objects as isolated units, \gom exposes multimodal models to relational data without requiring retraining or architectural changes. 
Through extensive evaluation, we show that \gom consistently enhances performance in spatially grounded tasks and offers a powerful, robust mechanism for structured visual understanding in lightweight open-source MLMs.
Our method outperforms competitive approaches while demonstrating the advantages of visual design over graph-verbalized text prompting.
Future directions include scene hypergraphs for complex scenes, stereo vision for improved depth reasoning, and temporal modeling for video understanding.
Furthermore, we project significant potential in the clinical domain, where \gom could be specialized and optimized for near-real-time augmentation.
This would enable medical MLMs to achieve superior test-time performance on tasks such as diagnostic classification and surgical video analysis.
As MLMs' capabilities continue to grow, we expect the benefits of \gom to amplify accordingly.


\section*{Acknowledgements}
Research partially supported by: AI-PACT (CUP B47H22004450008, B47H22004460001); National Plan PNC-I.1 DARE (PNC0000002, CUP B53C22006450001); PNRR Extended Partnership FAIR (PE00000013, Spoke 8); 2024 Scientific Research and High Technology Program, ``AI analysis for risk assessment of empty lymph nodes in endometrial cancer surgery,'' Fondazione Cassa di Risparmio in Bologna; Chips JU TRISTAN (G.A. 101095947). LG Solution Srl for co-funding L. Molfetta's PhD scholarship.


\clearpage
\bibliography{aaai2026_short}

\clearpage

\appendix

\section{\gom Algorithm Details}
\label{sec:appendix}

\paragraph{Algorithmic Summary of GoM.}  
We provide the complete algorithmic description of the Graph-of-Marks (\gom) method. The pipeline is structured modularly to reflect the main components described in Section~3, and is formalized through a sequence of pseudocode listings. 

\noindent\roundedtagWithSymbol{}{\textbf{Algorithm~\ref{alg:object-detection}}} (\textit{\gom: Object Detection \& Segmentation}) outlines the initial stage, in which a set of object detectors is applied to the input image $I$ to generate raw bounding boxes. These are subsequently merged using the Weighted Boxes Fusion (WBF) heuristic, filtered by a minimum confidence threshold, and refined into pixel-accurate object masks. This results in a set of region-level object candidates that serve as nodes in the scene graph.

\noindent\roundedtagWithSymbol{}{\textbf{Algorithm~\ref{alg:relation-estimation}}} (\textit{\gom: Spatial Relation Estimation}) computes directed edges between all object pairs based on three classes of spatial relations: directional (e.g., \texttt{above}, \texttt{right\_of}), depth-based (e.g., \texttt{in\_front\_of}, \texttt{behind}), and proximity-based (e.g., \texttt{near}), along with optional modifiers for closeness (e.g., \texttt{touching}). This module uses bounding box geometry and monocular depth maps to determine dominant spatial cues and populates a structured relation set $\mathcal{E}$ with labeled triplets of the form $(r_i, \text{relation}, r_j)$.

\noindent\roundedtagWithSymbol{}{\textbf{Algorithm~\ref{alg:query-filtering}}} (\textit{\gom: Query-based Graph Filtering}) performs prompt-driven pruning to reduce redundancy and focus the graph on query-relevant content. Objects are filtered incrementally through lexical match and semantic similarity (via cosine similarity over embeddings) with the user's textual query $T$. If multiple objects match, the graph is restricted to them and their pairwise relations. If a single object matches, only its outgoing edges and corresponding tail objects are retained. A fallback mechanism retains all nodes when no query-relevant matches are found. Subsequently, relation filtering is performed per head-object, selecting top-$k$ relations based on query relevance and spatial distance, and deduplicating inverse edges.

\noindent\roundedtagWithSymbol{}{\textbf{Algorithm~\ref{alg:scene-graph-rendering}}} (\textit{\gom: Scene Graph Rendering}) transforms the filtered object-relation graph into a pixel-level visual overlay. Object masks are rendered using class-specific colors and labeled with either numeric or textual IDs. Relations are displayed as curved arrows from head to tail, optionally annotated with their type (e.g., \texttt{left\_of}, \texttt{near}). To avoid visual occlusions and maximize interpretability, a placement strategy dynamically adjusts the location of ID boxes and relation labels, applies dashed guide lines to link displaced elements to their referents, and resolves overlaps incrementally through axis-aligned displacement.

\noindent\roundedtagWithSymbol{}{\textbf{Algorithm~\ref{alg:gom-pipeline}}} (\textit{\gom: Main Pipeline for MLM Scene Graph Prompting}) summarizes the full end-to-end flow, combining all components into a unified prompting strategy. Given image $I$ and prompt $T$, it produces a scene-graph-rendered image $I^{SG}$ and an optional verbalized scene graph text $T_{SG}$, which are passed to the multimodal language model. The output is a probability distribution over text completions, conditioned on both visual and textual input. This main procedure provides a concise yet extensible interface for integrating \gom with any frozen vision-language backbone.

Each algorithm operates with configurable threshold parameters ($\tau$ values) that can be tuned for specific datasets or task requirements.

\begin{algorithm}
\caption{\gom: Object Detection \& Segmentation}
\label{alg:object-detection}
\begin{algorithmic}[1]
\Require Image $I$, detector ensemble $\mathcal{D}$
\Ensure Object regions $R$
\State $\mathcal{B} \leftarrow \emptyset$
\For{$d \in \mathcal{D}$}
    \State $\mathcal{B} \leftarrow \mathcal{B} \cup \texttt{Detect}(I, d)$
\EndFor
\State $\mathcal{B}_{fused} \leftarrow \text{WBF}(\mathcal{B}, \tau_\text{OD-min-conf})$
\State $R \leftarrow \texttt{Segment}(\mathcal{B}_{fused}, I)$
\State \Return $R$
\end{algorithmic}
\end{algorithm}

\begin{algorithm}
\caption{\gom: Spatial Relation Estimation}
\label{alg:relation-estimation}
\begin{algorithmic}[1]
\Require Object regions $R$, image $I$
\Ensure Relations $\mathcal{E}$
\State $\mathcal{E} \leftarrow \emptyset$
\For{$(r_i, r_j) \in R \times R, i \neq j$}
    \State $rel \leftarrow \emptyset$
    \State $(cx_i, cy_i), (cx_j, cy_j) \leftarrow \text{Centers}(r_i, r_j)$
    \State $(dx, dy) \leftarrow (cx_j - cx_i, cy_j - cy_i)$
    
    \State \textit{// Directional}
    \If{$|dy| \geq |dx|$ and $|dy| > \tau_\text{dir-margin}$}
        \State $rel \leftarrow rel \cup \texttt{aboveORbelow}(\texttt{sign}(dy))$
    \ElsIf{$|dx| > \tau_{margin}$}
        \State $rel \leftarrow rel \cup \texttt{leftORright}(\texttt{sign}(dx))$
    \EndIf
    
    \State \textit{// Depth}
    \State $\delta_i, \delta_j \leftarrow \text{Depth}(I, cx_i, cx_j)$
    \If{$|\delta_j - \delta_i| > \tau_\text{z-diff}$}
        \State $rel \leftarrow rel \cup \texttt{frontORbehind}(\texttt{sign}(\delta_j - \delta_i))$
    \EndIf
    
    \State \textit{// Proximity \& Modifiers}
    \If{$|rel| = 0$ and $|(cx_i, cy_i)-(cx_j,cy_j)|_2 < \tau_\text{near}$}
        \State $rel \leftarrow rel \cup \texttt{near}$
    \EndIf

    \For{${rel}_k \in rel$}
        \State $\mathcal{E} \leftarrow \mathcal{E} \cup \{(r_i, {rel}_k, r_j)\}$
    \EndFor
\EndFor
\State \Return $\mathcal{E}$
\end{algorithmic}
\end{algorithm}

\begin{algorithm}
\caption{\gom: Query-based Graph Filtering}
\label{alg:query-filtering}
\begin{algorithmic}[1]
\Require Objects $\mathcal{O}$, Relations $\mathcal{E}$, query prompt $T$
\Ensure Filtered objects $\mathcal{O}_T$, filtered relations $\mathcal{E}_T$
\State $\mathcal{O}_T \leftarrow \emptyset$, $\mathcal{E}_T \leftarrow \emptyset$

\State \textit{// Object Filtering}
\For{$o \in \mathcal{O}$}
    \State $L_o \leftarrow \texttt{LabelsAndAliases}(o)$
    \If{$\texttt{LexMatch}(T, L_o)$}
        \State $\mathcal{O}_T \leftarrow \mathcal{O}_T \cup \{o\}$
    \ElsIf{$\texttt{cos\_sim}(T, L_o) > \tau_\text{obj}$}
        \State $\mathcal{O}_T \leftarrow \mathcal{O}_T \cup \{o\}$
    \EndIf
\EndFor

\If{$|\mathcal{O}_T| = 0$}
    \State $\mathcal{O}_T \leftarrow \mathcal{O}$ \Comment{Fallback for low specificity}
\ElsIf{$|\mathcal{O}_T| = 1$}
    \State $o \leftarrow \text{only element in } \mathcal{O}_T$
    \State $\mathcal{E}_T \leftarrow \{e \in \mathcal{E} \mid \text{Head}(e) = o\}$
    \State $\mathcal{O}_T \leftarrow \mathcal{O}_T \cup \{\text{Tail}(e) \mid e \in \mathcal{E}_T\}$
\Else
    \State $\mathcal{E}_T \leftarrow \{e \in \mathcal{E} \mid \text{Head}(e), \text{Tail}(e) \in \mathcal{O}_T\}$
\EndIf

\State \textit{// Relation Filtering}
\State $\mathcal{E}_T^{final} \leftarrow \emptyset$
\For{$o \in \mathcal{O}_T$}
    \State $\mathcal{E}_o \leftarrow \{e \in \mathcal{E}_T \mid \text{Head}(e) = o\}$
    \For{$e \in \mathcal{E}_o$}
        \State $r_e \leftarrow \texttt{RelLabel}(e)$
        \State $Q_r \leftarrow \texttt{ExtractRelTerms}(T)$
        \If{$\texttt{LexMatch}(r_e, Q_r)$}
            \State $s(e) \leftarrow 0$ \Comment{0 = relevant}
        \ElsIf{$\texttt{cos\_sim}(r_e, Q_r) > \tau_\text{rel}$}
            \State $s(e) \leftarrow 0$
        \Else
            \State $s(e) \leftarrow 1$
        \EndIf
        \State $d(e) \leftarrow \texttt{SpatialDist}(\text{Head}(e), \text{Tail}(e))$
    \EndFor
    \State \texttt{Sort} $\mathcal{E}_o$ by $(s(e), d(e))$
    \State $\mathcal{E}_T^{final} \leftarrow \mathcal{E}_T^{final} \cup \texttt{TopK}(\mathcal{E}_o)$
\EndFor

\State $\mathcal{E}_T \leftarrow \texttt{Deduplicate}(\mathcal{E}_T^{final})$
\State \Return $(\mathcal{O}_T, \mathcal{E}_T)$
\end{algorithmic}
\end{algorithm}

\begin{algorithm}
\caption{\gom: Scene Graph Rendering}
\label{alg:scene-graph-rendering}
\begin{algorithmic}[1]
\Require Image $I$, filtered objects $\mathcal{O}_T$, filtered relations $\mathcal{E}_T$
\Ensure Rendered image with scene graph $I_{SG}$

\State $I_{SG} \leftarrow I$

\State \textit{// Node Marks: Object masks and IDs}
\For{$o \in \mathcal{O}_T$}
    \State $c \leftarrow \texttt{ClassColor}(\texttt{Label}(o))$
    \State $mask(o) \leftarrow \texttt{DrawMask}(I_{SG}, o, \texttt{color}=c)$
    \State $id(o) \leftarrow \texttt{AssignID}(o)$
    \State $p_{id} \leftarrow \texttt{PlaceID}(o, id(o), \texttt{color}=c)$
\EndFor

\State \textit{// Edge Marks: Arrows and labels}
\For{$e = (h, r, t) \in \mathcal{E}_T$}
    \State $c_h \leftarrow \texttt{ClassColor}(\texttt{Label}(h))$
    \State $a \leftarrow \texttt{DrawArrow}(I_{SG}, h, t, \texttt{color}=c_h)$
    \If{\texttt{RenderLabels}}
        \State $p_r \leftarrow \texttt{PlaceLabel}(h, t, r, \texttt{color}=c_h)$
    \EndIf
\EndFor

\State \textit{// Mark Allocation Strategy}
\State $\mathcal{M} \leftarrow$ all ID and relation marks
\State \texttt{ResolveOverlaps}$(\mathcal{M}, I_{SG})$
\State \texttt{AttachDashedGuides}$(\mathcal{M}, I_{SG})$
\State \Return $I_{SG}$
\end{algorithmic}
\end{algorithm}

\begin{algorithm}
\caption{\gom: Main Pipeline for MLM Scene Graph Prompting}
\label{alg:gom-pipeline}
\begin{algorithmic}[1]
\Require Image $I$, text prompt $T$
\Ensure MLM output distribution $P_{MLM}(\cdot | I^{SG}, T_{SG})$

\State $\mathcal{O} \leftarrow \texttt{ObjectDetection}(I)$ \Comment{Alg.~\ref{alg:object-detection}}
\State $\mathcal{E} \leftarrow \texttt{RelationEstimation}(\mathcal{O}, I)$ \Comment{Alg.~\ref{alg:relation-estimation}}
\State $\mathcal{O}_T, \mathcal{E}_T \leftarrow \texttt{QueryFiltering}(\mathcal{O}, \mathcal{E}, T)$ \Comment{Alg.~\ref{alg:query-filtering}}
\State $I^{SG} \leftarrow \texttt{SceneGraphRend}(I, \mathcal{O}_T, \mathcal{E}_T)$ \Comment{Alg.~\ref{alg:scene-graph-rendering}}

\If{SG-grounded mode}
    \State $T_{SG} \leftarrow T + \texttt{Verbalize}(\mathcal{O}_T, \mathcal{E}_T)$
\Else
    \State $T_{SG} \leftarrow T$
\EndIf

\State \Return $P_{MLM}(\cdot \mid I^{SG}, T_{SG})$
\end{algorithmic}
\end{algorithm}

\section{Prompt Templates}

\begin{figure}[th!]
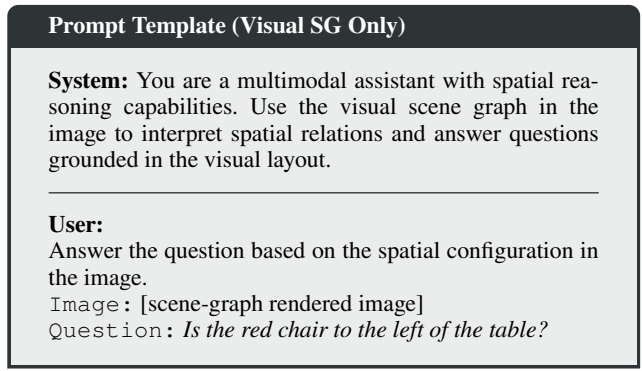

\centering  
\small
\begin{tcolorbox}[colback=gray!10!white, colframe=gray!80!black, title=\textbf{Prompt Template (Visual SG Only)}, fonttitle=\bfseries, sharp corners=south]
\textbf{System:} You are a multimodal assistant with spatial reasoning capabilities. Use the visual scene graph in the image to interpret spatial relations and answer questions grounded in the visual layout.

\rule{\linewidth}{0.4pt}

\vspace{0.5em}
\textbf{User:}\\
Answer the question based on the spatial configuration in the image.

\texttt{Image:} [scene-graph rendered image]

\texttt{Question:} \emph{Is the red chair to the left of the table?}
\end{tcolorbox}
\caption{\faFilePictureO\,\gom prompt template used in the \textsc{Visual SG Only} condition.}
\label{fig:prompt-visual-only}
\end{figure}

\begin{figure}[th!]
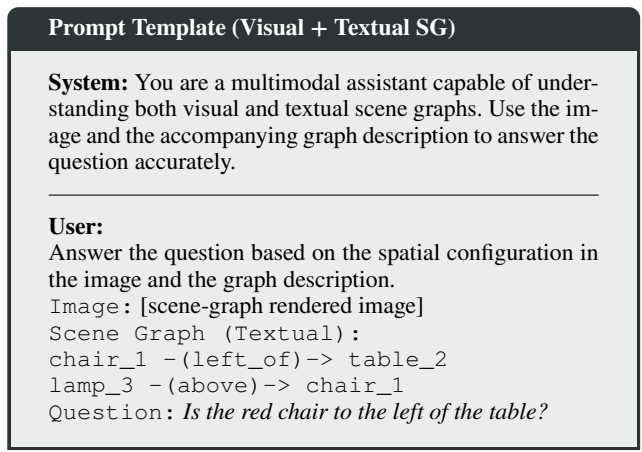

\centering  
\small
\begin{tcolorbox}[colback=gray!10!white, colframe=gray!80!black, title=\textbf{Prompt Template (Visual + Textual SG)}, fonttitle=\bfseries, sharp corners=south]
\textbf{System:} You are a multimodal assistant capable of understanding both visual and textual scene graphs. Use the image and the accompanying graph description to answer the question accurately.

\rule{\linewidth}{0.4pt}

\vspace{0.5em}
\textbf{User:}\\
Answer the question based on the spatial configuration in the image and the graph description.

\texttt{Image:} [scene-graph rendered image]

\texttt{Scene Graph (Textual):}\\
\texttt{chair\_1 --(left\_of)--> table\_2}\\
\texttt{lamp\_3 --(above)--> chair\_1}

\texttt{Question:} \emph{Is the red chair to the left of the table?}
\end{tcolorbox}
\caption{\faFilePictureO\faFileTextO\,\gom prompt template used in the \textsc{Visual + Textual SG} condition.}
\label{fig:prompt-visual-textual}
\end{figure}

To support reproducibility, we include here the exact prompt formats used during inference. \textbf{Figure~\ref{fig:prompt-visual-only}} shows the template used for the \textsc{Visual SG Only} condition, where the model is prompted solely with the scene-graph-rendered image and the natural language question. \textbf{Figure~\ref{fig:prompt-visual-textual}} presents the template for the \textsc{Visual + Textual SG} setting, which augments the image with a structured textual rendering of the filtered scene graph. The textual graph follows a canonical format of subject–relation–object triplets with disambiguated entity IDs, and is intended to facilitate parsing by language-pretrained models. Both prompt formats are designed to be minimal and task-general, enabling zero-shot inference with off-the-shelf MLMs and facilitating a fair comparison with our methods to assess \gom's contribution clearly and without bias from external factors.

\section{Qualitative Examples}
\begin{figure*}[ht!]
\centering

\textbf{Target object instances:}\\

[ \textit{\quotes{a cloth sofa with a towel on it}}, \textit{\quotes{a brown couch to the left of an easy chair in the front of a fireplace}} ]

\vspace{0.2cm}

\hspace{5.95cm}
\begin{subfigure}[t]{0.32\textwidth}
    \centering
    \fontsize{8}{11}\selectfont
    \includegraphics[width=\textwidth]{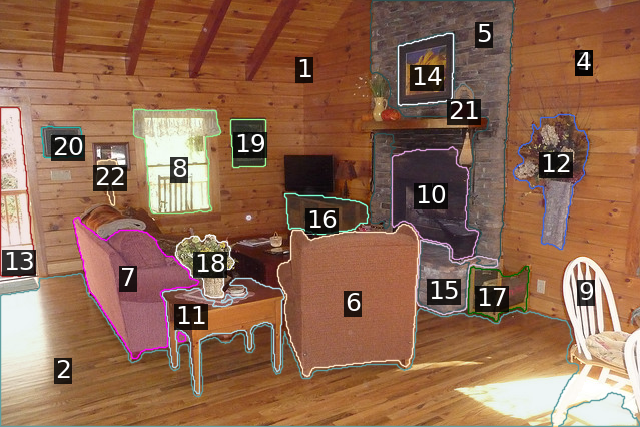}\\[-6pt]
    \colorbox{baseline}{%
    \begin{minipage}[t][2.8cm][t]{.95\textwidth}
        \vspace{2pt}
        {\arrayrulecolor{tablebordercolor} 
        \renewcommand{\arraystretch}{1.5} 
        \begin{tabularx}{\textwidth}{XX} 
        \multicolumn{1}{>{\arraybackslash}X}{\fontsize{9}{11}\selectfont$\textbf{Method:}\;\,\segNumObj[0.15in]$} & \multicolumn{1}{>{\arraybackslash}X}{\fontsize{9}{11}\selectfont$\textbf{Accurate:}\;\,\textcolor{error}{\text{\faClose}}$} \\[-1mm] 
        \multicolumn{2}{>{\hsize=\dimexpr2\hsize+2\tabcolsep+\arrayrulewidth\relax}X}{The ID for the cloth sofa with a towel on it is \textcolor{error}{18}, and the ID for the brown couch to the left of an easy chair in front of the fireplace is \textcolor{error}{6}.} \\
        \end{tabularx}}
        \vspace{2pt}
        \end{minipage}
    }
\end{subfigure}
\hfill
\begin{subfigure}[t]{0.32\textwidth}
    \centering
    \fontsize{8}{11}\selectfont
    \includegraphics[width=\textwidth]{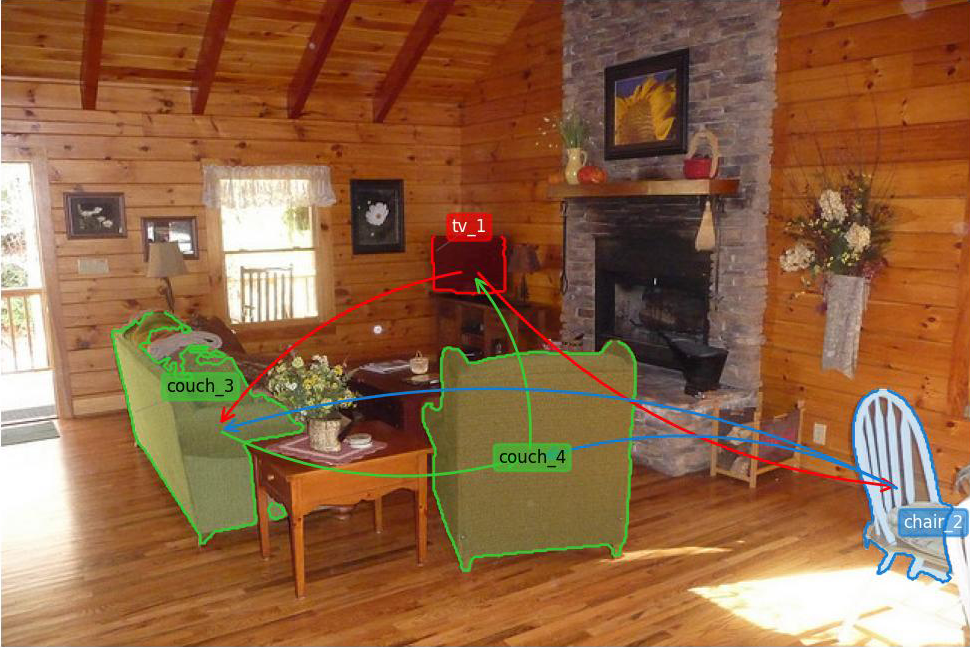}\\[-6pt]
    \colorbox{gom}{%
    \begin{minipage}[t][2.8cm][t]{.95\textwidth}
        \vspace{2pt}
        {\arrayrulecolor{tablebordercolor} 
        \renewcommand{\arraystretch}{1.5} 
        \begin{tabularx}{\textwidth}{XX} 
        \multicolumn{1}{>{\arraybackslash}X}{\fontsize{9}{11}\selectfont$\textbf{Method:}\;\,\gomLabObj[0.35in]$} & \multicolumn{1}{>{\arraybackslash}X}{\fontsize{9}{11}\selectfont$\textbf{Accurate:}\;\,\textcolor{correct}{\text{\faCheck}}$} \\[-1mm] 
        \multicolumn{2}{>{\hsize=\dimexpr2\hsize+2\tabcolsep+\arrayrulewidth\relax}X}{The IDs for the objects you've described are:\newline - **A cloth sofa with a towel on it**: The ID is `couch\_4'\newline - **A brown couch to the left of an easy chair in front of a fireplace**: The ID is `couch\_3'} \\
        \end{tabularx}}
        \vspace{2pt}
        \end{minipage}
    }  
\end{subfigure}

\vspace{0.1cm}

\begin{subfigure}[t]{0.32\textwidth}
    \centering
    \fontsize{8}{11}\selectfont
    \includegraphics[width=\textwidth]{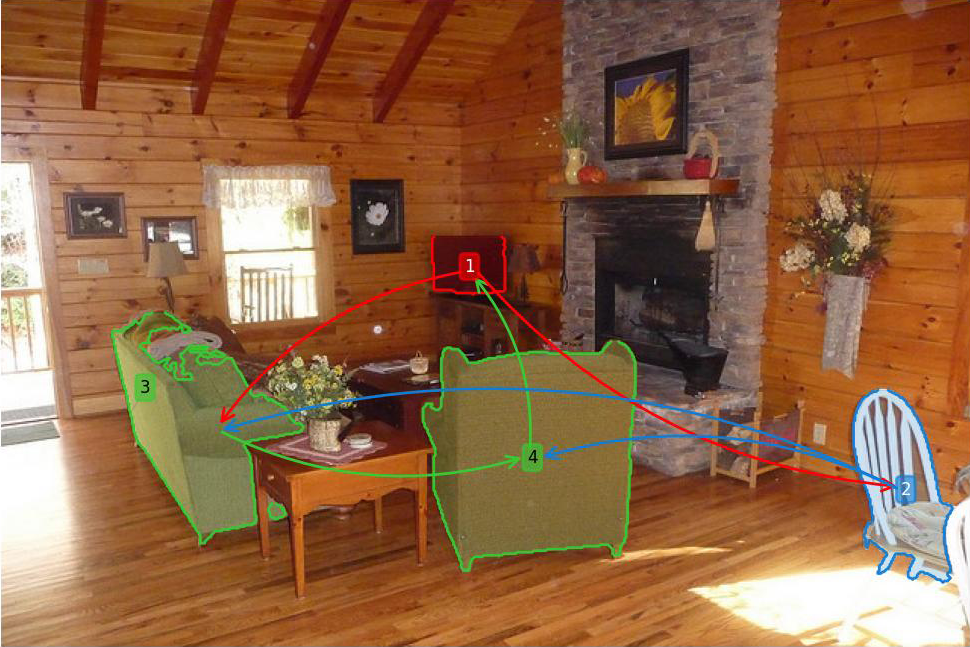}\\[-6pt]
    \colorbox{gom}{%
    \begin{minipage}[t][3.1cm][t]{.95\textwidth}
        \vspace{2pt}
        {\arrayrulecolor{tablebordercolor} 
        \renewcommand{\arraystretch}{1.5} 
        \begin{tabularx}{\textwidth}{XX} 
        \multicolumn{1}{>{\arraybackslash}X}{\fontsize{9}{11}\selectfont$\textbf{Method:}\;\,\gomNumObj[0.35in]$} & \multicolumn{1}{>{\arraybackslash}X}{\fontsize{9}{11}\selectfont$\textbf{Accurate:}\;\,\textcolor{correct}{\text{\faCheck}}$} \\[-1mm] 
        \multicolumn{2}{>{\hsize=\dimexpr2\hsize+2\tabcolsep+\arrayrulewidth\relax}X}{The IDs for the objects you've described are:\newline 1. The cloth sofa with a towel on it is labeled as **4**.\newline 2. The brown couch to the left of an easy chair in front of a fireplace is labeled as **3**.} \\
        \end{tabularx}}
        \vspace{2pt}
        \end{minipage}
    }
\end{subfigure}
\hfill
\begin{subfigure}[t]{0.32\textwidth}
    \centering
    \fontsize{8}{11}\selectfont
    \includegraphics[width=\textwidth]{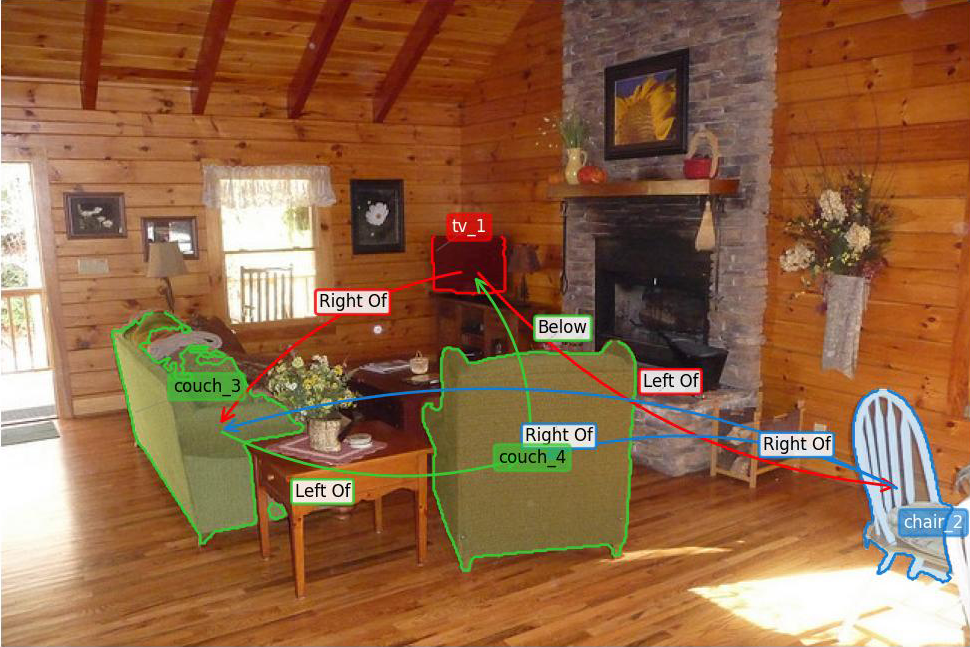}\\[-6pt]
    \colorbox{gom}{%
    \begin{minipage}[t][3.1cm][t]{.95\textwidth}
        \vspace{2pt}
        {\arrayrulecolor{tablebordercolor} 
        \renewcommand{\arraystretch}{1.5} 
        \begin{tabularx}{\textwidth}{XX} 
        \multicolumn{1}{>{\arraybackslash}X}{\fontsize{9}{11}\selectfont$\textbf{Method:}\;\,\gomLabObjLabRel[0.35in]$} & \multicolumn{1}{>{\arraybackslash}X}{\fontsize{9}{11}\selectfont$\textbf{Accurate:}\;\,\textcolor{error}{\text{\faClose}}$} \\[-1mm] 
        \multicolumn{2}{>{\hsize=\dimexpr2\hsize+2\tabcolsep+\arrayrulewidth\relax}X}{The IDs for the objects you've described are:\newline - **Cloth sofa with a towel on it**: The ID is \textcolor{error}{`couch\_3'}\newline - **Brown couch to the left of an easy chair in front of a fireplace**: The ID is \textcolor{error}{`couch\_4'}} \\
        \end{tabularx}}
        \vspace{2pt}
        \end{minipage}
    }
\end{subfigure}
\hfill
\begin{subfigure}[t]{0.32\textwidth}
    \centering
    \fontsize{8}{11}\selectfont
    \includegraphics[width=\textwidth]{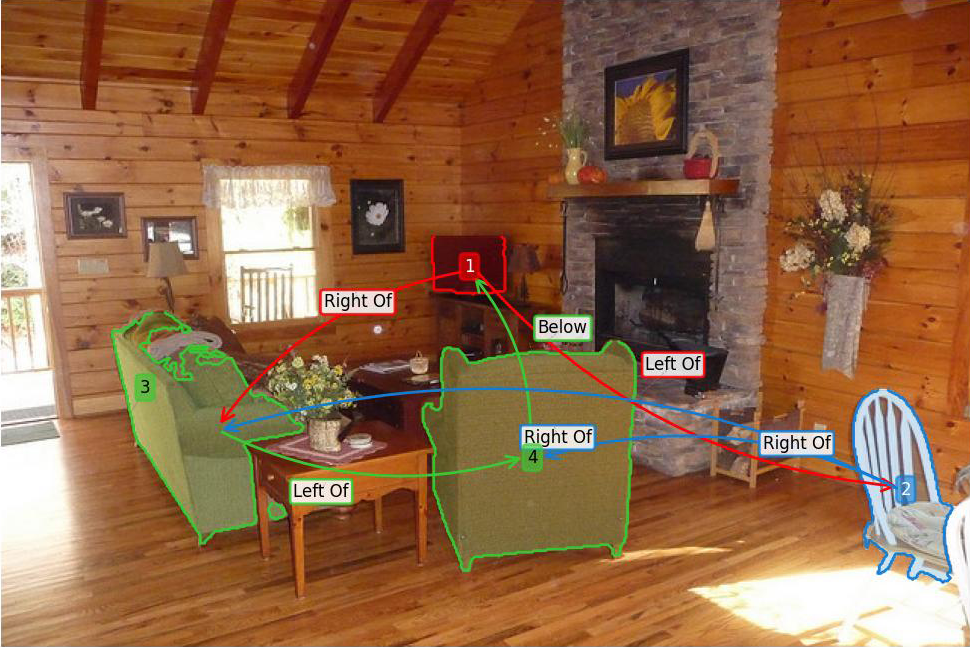}\\[-6pt]
    \colorbox{gom}{%
    \begin{minipage}[t][3.1cm][t]{.95\textwidth}
        \vspace{2pt}
        {\arrayrulecolor{tablebordercolor} 
        \renewcommand{\arraystretch}{1.5} 
        \begin{tabularx}{\textwidth}{XX} 
        \multicolumn{1}{>{\arraybackslash}X}{\fontsize{9}{11}\selectfont$\textbf{Method:}\;\,\gomNumObjLabRel[0.35in]$} & \multicolumn{1}{>{\arraybackslash}X}{\fontsize{9}{11}\selectfont$\textbf{Accurate:}\;\,\textcolor{correct}{\text{\faCheck}}$} \\[-1mm] 
        \multicolumn{2}{>{\hsize=\dimexpr2\hsize+2\tabcolsep+\arrayrulewidth\relax}X}{The IDs for the objects you've described are:\newline 1. The cloth sofa with a towel on it is labeled as **ID 4**.\newline 2. The brown couch to the left of an easy chair in front of a fireplace is labeled as **ID 3**.} \\
        \end{tabularx}}
        \vspace{2pt}
        \end{minipage}
    }
\end{subfigure}

\caption{\textbf{Qualitative example illustrating the impact of image preprocessing on REF performance.} The same question from RefCOCOg is posed to Qwen2.5-7B using 5 different hard visual prompts, highlighting how pixel transformations can influence the model's responses. Gray boxes denote baseline outputs, while blue boxes indicate those from our proposed \gom. The original font size and line thickness values are preserved. Best viewed when zoomed in.}
\label{fig:refcocog_example}
\end{figure*}

\begin{figure*}[ht!]
\centering

\textbf{Question:} \textit{\quotes{Is the bowl in the top part or in the bottom of the picture?}}

\vspace{0.2cm}

\begin{subfigure}[t]{0.32\textwidth}
    \centering
    \fontsize{8}{11}\selectfont
    \includegraphics[width=\textwidth]{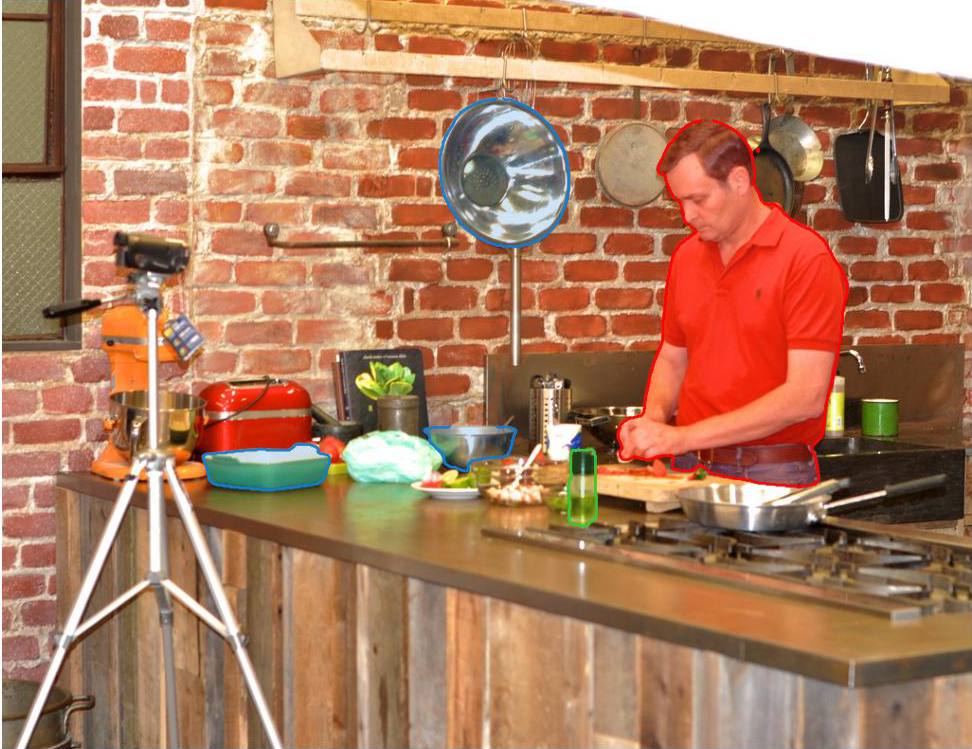}\\[-6pt]
    \colorbox{baseline}{%
    \begin{minipage}[t][1.8cm][t]{.95\textwidth}
        \vspace{2pt}
        {\arrayrulecolor{tablebordercolor} 
        \renewcommand{\arraystretch}{1.5} 
        \begin{tabularx}{\textwidth}{XX} 
        \multicolumn{1}{>{\arraybackslash}X}{\fontsize{9}{11}\selectfont$\textbf{Method:}\;\,\seg[0.15in]$} 
        & 
        \multicolumn{1}{>{\arraybackslash}X}{\fontsize{9}{11}\selectfont$\textbf{Accurate:}\;\,\textcolor{error}{\text{\faClose}}$} 
        \\[-1mm]
        \multicolumn{2}{>{\hsize=\dimexpr2\hsize+2\tabcolsep+\arrayrulewidth\relax}X}{The bowl is in the \textcolor{error}{bottom} part of the picture.} \\
        \end{tabularx}}
        \vspace{2pt}
        \end{minipage}
    }
\end{subfigure}
\hfill
\begin{subfigure}[t]{0.32\textwidth}
    \centering
    \fontsize{8}{11}\selectfont
    \includegraphics[width=\textwidth]{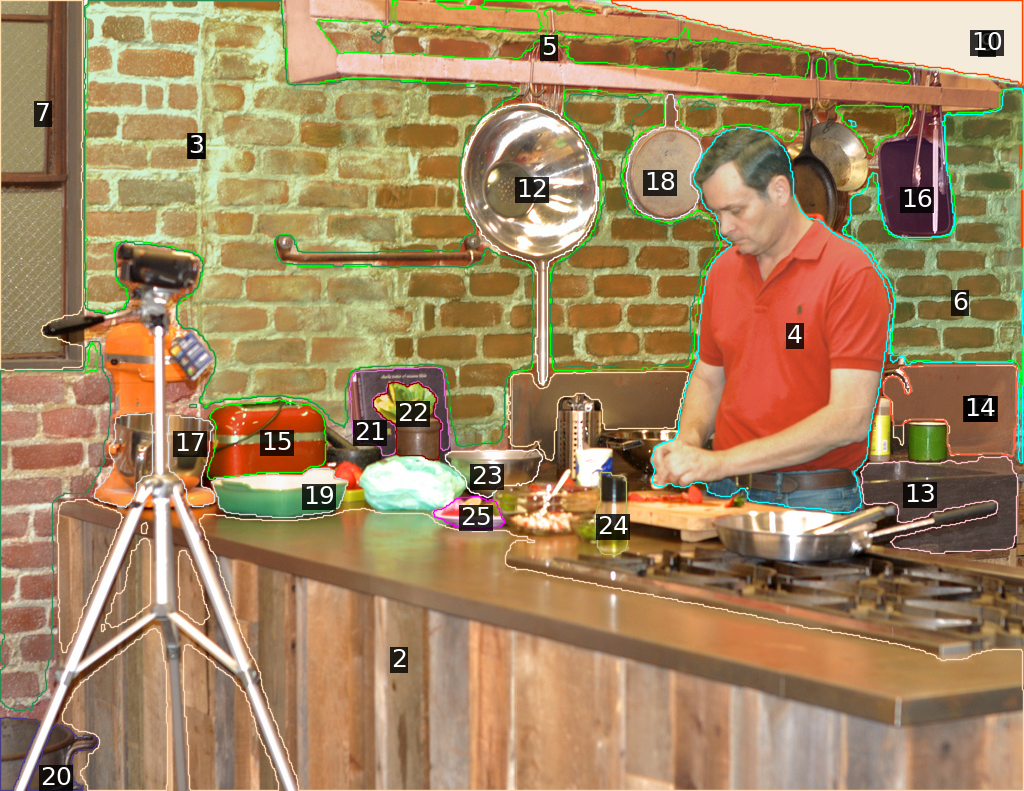}\\[-6pt]
    \colorbox{baseline}{%
    \begin{minipage}[t][1.8cm][t]{.95\textwidth}
        \vspace{2pt}
        {\arrayrulecolor{tablebordercolor} 
        \renewcommand{\arraystretch}{1.5} 
        \begin{tabularx}{\textwidth}{XX} 
        \multicolumn{1}{>{\arraybackslash}X}{\fontsize{9}{11}\selectfont$\textbf{Method:}\;\,\segNumObj[0.15in]$} & \multicolumn{1}{>{\arraybackslash}X}{\fontsize{9}{11}\selectfont$\textbf{Accurate:}\;\,\textcolor{error}{\text{\faClose}}$} \\[-1mm] 
        \multicolumn{2}{>{\hsize=\dimexpr2\hsize+2\tabcolsep+\arrayrulewidth\relax}X}{The bowl is located in the  \textcolor{error}{bottom} part of the picture.} \\
        \end{tabularx}}
        \vspace{2pt}
        \end{minipage}
    }
\end{subfigure}
\hfill
\begin{subfigure}[t]{0.32\textwidth}
    \centering
    \fontsize{8}{11}\selectfont
    \includegraphics[width=\textwidth]{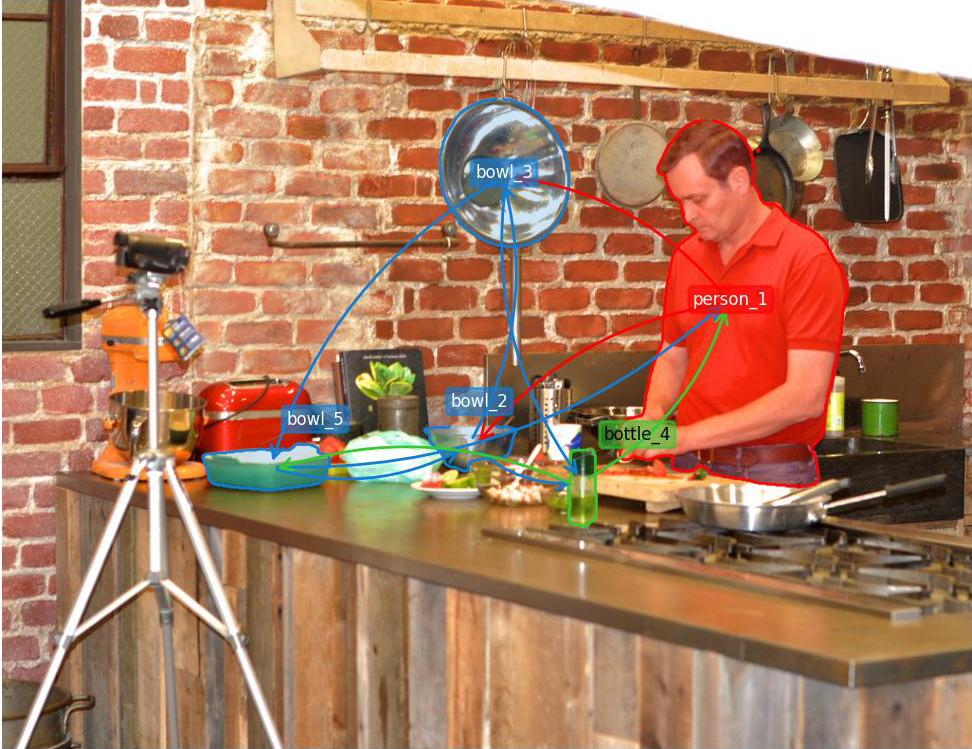}\\[-6pt]
    \colorbox{gom}{%
    \begin{minipage}[t][1.8cm][t]{.95\textwidth}
        \vspace{2pt}
        {\arrayrulecolor{tablebordercolor} 
        \renewcommand{\arraystretch}{1.5} 
        \begin{tabularx}{\textwidth}{XX} 
        \multicolumn{1}{>{\arraybackslash}X}{\fontsize{9}{11}\selectfont$\textbf{Method:}\;\,\gomLabObj[0.35in]$} & \multicolumn{1}{>{\arraybackslash}X}{\fontsize{9}{11}\selectfont$\textbf{Accurate:}\;\,\textcolor{correct}{\text{\faCheck}}$} \\[-1mm] 
        \multicolumn{2}{>{\hsize=\dimexpr2\hsize+2\tabcolsep+\arrayrulewidth\relax}X}{The bowl is in the top part of the picture.} \\
        \end{tabularx}}
        \vspace{2pt}
        \end{minipage}
    }  
\end{subfigure}

\vspace{0.1cm}

\begin{subfigure}[t]{0.32\textwidth}
    \centering
    \fontsize{8}{11}\selectfont
    \includegraphics[width=\textwidth]{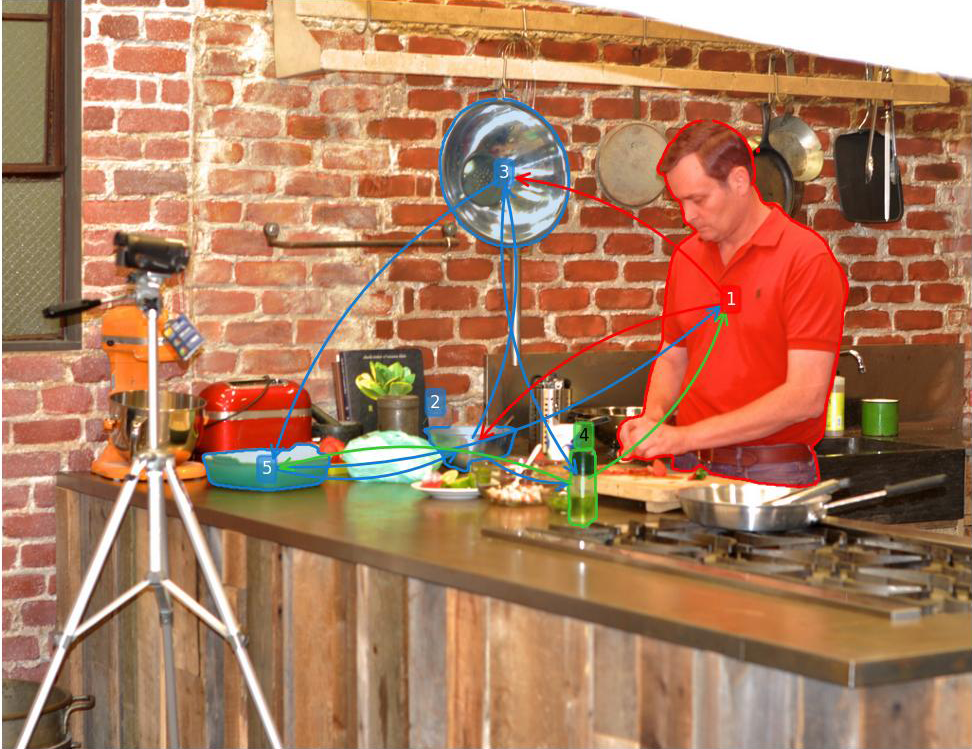}\\[-6pt]
    \colorbox{gom}{%
    \begin{minipage}[t][1.7cm][t]{.95\textwidth}
        \vspace{2pt}
        {\arrayrulecolor{tablebordercolor} 
        \renewcommand{\arraystretch}{1.5} 
        \begin{tabularx}{\textwidth}{XX} 
        \multicolumn{1}{>{\arraybackslash}X}{\fontsize{9}{11}\selectfont$\textbf{Method:}\;\,\gomNumObj[0.35in]$} & \multicolumn{1}{>{\arraybackslash}X}{\fontsize{9}{11}\selectfont$\textbf{Accurate:}\;\,\textcolor{correct}{\text{\faCheck}}$} \\[-1mm] 
        \multicolumn{2}{>{\hsize=\dimexpr2\hsize+2\tabcolsep+\arrayrulewidth\relax}X}{The bowl is in the top part.} \\
        \end{tabularx}}
        \vspace{2pt}
        \end{minipage}
    }
\end{subfigure}
\hfill
\begin{subfigure}[t]{0.32\textwidth}
    \centering
    \fontsize{8}{11}\selectfont
    \includegraphics[width=\textwidth]{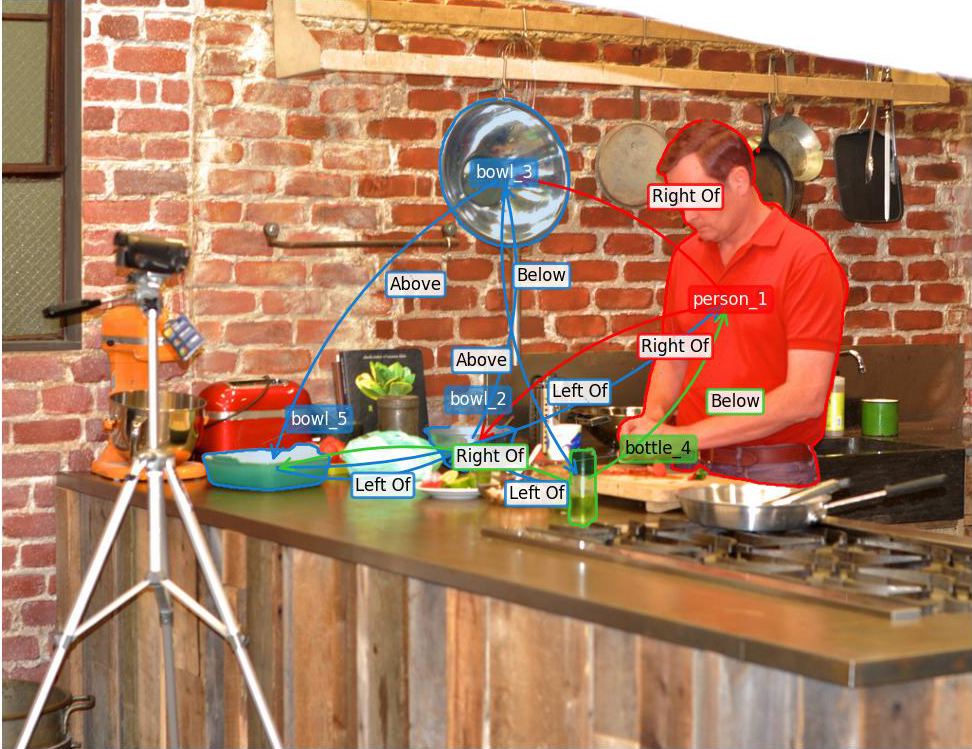}\\[-6pt]
    \colorbox{gom}{%
    \begin{minipage}[t][1.7cm][t]{.95\textwidth}
        \vspace{2pt}
        {\arrayrulecolor{tablebordercolor} 
        \renewcommand{\arraystretch}{1.5} 
        \begin{tabularx}{\textwidth}{XX} 
        \multicolumn{1}{>{\arraybackslash}X}{\fontsize{9}{11}\selectfont$\textbf{Method:}\;\,\gomLabObjLabRel[0.35in]$} & \multicolumn{1}{>{\arraybackslash}X}{\fontsize{9}{11}\selectfont$\textbf{Accurate:}\;\,\textcolor{correct}{\text{\faCheck}}$} \\[-1mm] 
        \multicolumn{2}{>{\hsize=\dimexpr2\hsize+2\tabcolsep+\arrayrulewidth\relax}X}{The bowl is in the top part of the image.} \\
        \end{tabularx}}
        \vspace{2pt}
        \end{minipage}
    }
\end{subfigure}
\hfill
\begin{subfigure}[t]{0.32\textwidth}
    \centering
    \fontsize{8}{11}\selectfont
    \includegraphics[width=\textwidth]{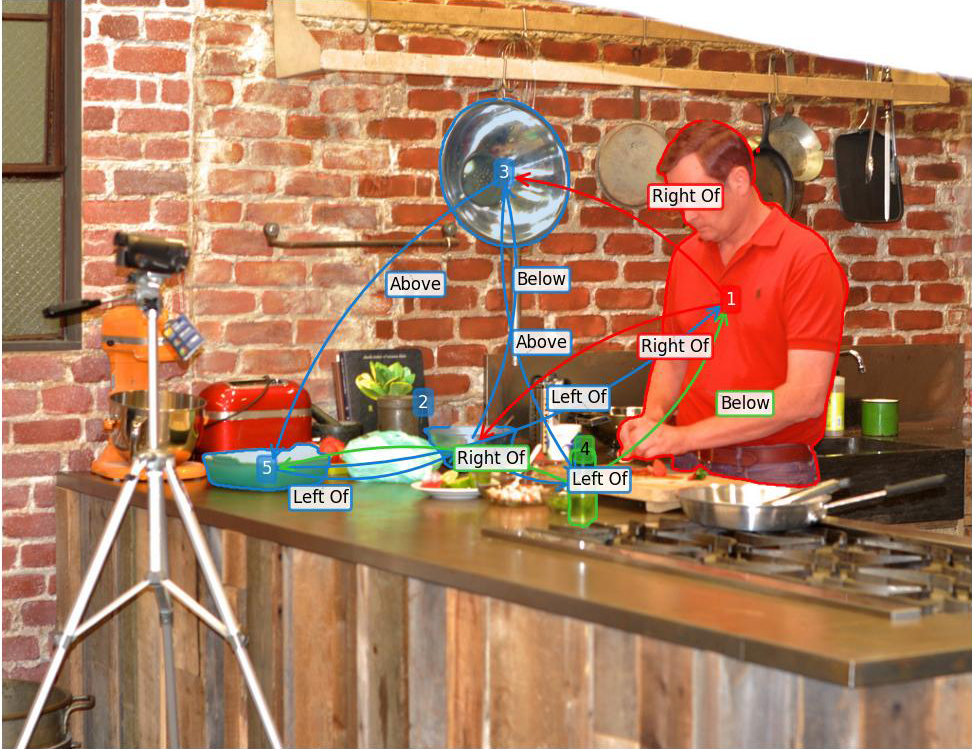}\\[-6pt]
    \colorbox{gom}{%
    \begin{minipage}[t][1.7cm][t]{.95\textwidth}
        \vspace{2pt}
        {\arrayrulecolor{tablebordercolor} 
        \renewcommand{\arraystretch}{1.5} 
        \begin{tabularx}{\textwidth}{XX} 
        \multicolumn{1}{>{\arraybackslash}X}{\fontsize{9}{11}\selectfont$\textbf{Method:}\;\,\gomNumObjLabRel[0.35in]$} & \multicolumn{1}{>{\arraybackslash}X}{\fontsize{9}{11}\selectfont$\textbf{Accurate:}\;\,\textcolor{correct}{\text{\faCheck}}$} \\[-1mm] 
        \multicolumn{2}{>{\hsize=\dimexpr2\hsize+2\tabcolsep+\arrayrulewidth\relax}X}{The bowl is in the top part of the picture.} \\
        \end{tabularx}}
        \vspace{2pt}
        \end{minipage}
    }
\end{subfigure}

\caption{\textbf{Additional qualitative example illustrating the impact of image preprocessing on VQA performance.} The same question from GQA is posed to Qwen2.5-7B using 5 different hard visual prompts, highlighting how pixel transformations can influence the model's responses. Gray boxes denote baseline outputs, while blue boxes indicate those from our proposed \gom. The original font size and line thickness values are preserved. Best viewed when zoomed in.}
\label{fig:gqa_example}
\end{figure*}

We present additional qualitative examples to illustrate the impact of structured scene rendering on spatial understanding in VLMs. 
In Figure~\ref{fig:refcocog_example}, we consider a sample from the RefCOCOg dataset and apply the same spatial reasoning question to multiple visual prompts, including the SoM baseline and four variants of our \gom method. 
The example highlights how simple segmentation overlays can lead to misinterpretations due to ambiguous spatial placement or lack of relation anchoring. 
In contrast, we observed that \gom correctly resolves the query.
However, we noticed that textual labels occasionally cause \gom to fail on REC tasks, likely because textual features are unhelpful and add noise given the task requirements.
Figure~\ref{fig:gqa_example} replicates the same analysis on a VQA-style question from GQA, again showing that models are sensitive to small visual design changes. 
In both cases, relation-labeled \gom variants tend to yield the most precise and interpretable outputs.
These examples visually reinforce the central claim that structured visual rendering can substantially improve spatial reasoning without fine-tuning the underlying model.

\end{document}

%% file: custom_imports.tex
\usepackage{comment}

\usepackage{amsmath}
\usepackage{amssymb} 
\usepackage{textcomp} 
\usepackage{verbatim}
\usepackage{pifont} 
\usepackage{fontawesome} 
\usepackage[clock]{ifsym} 
\usepackage[notextcomp]{stix} 
\usepackage{upgreek} 
\usepackage{xspace} 
\usepackage{upquote}
\usepackage{bbding}

\usepackage{soul} 
\usepackage{transparent}

\usepackage[most]{tcolorbox} 

\usepackage{adjustbox}

\usepackage{xparse}
\usepackage{scalerel} 
\usepackage{pgfplots}
\usepackage{pgfplotstable}
\usepgfplotslibrary{groupplots}
\usepgfplotslibrary{statistics}
\usetikzlibrary{matrix} 
\pgfplotsset{compat=newest}
\usetikzlibrary{pgfplots.fillbetween,patterns} 
\usepackage{tikz}
\usepgfplotslibrary{patchplots} 
\usepackage{caption}
\usepackage{subcaption}

\usepackage{array}
\usepackage{multirow}
\usepackage{colortbl}
\usepackage{arydshln} 
\usepackage{tabularx} 
\usepackage[para,online,flushleft]{threeparttable} 
\usepackage{stackengine}
\usepackage{hhline} 
\usepackage{diagbox} 
\usepackage{booktabs}
\usepackage{nicematrix}
\usepackage{bm}

\usepackage{makecell}
\usepackage{bm} 
\usepackage{amsfonts} 

\usepackage{mathtools}

\usepackage{enumitem} 

\usepackage{pdfpages}

\usepackage{lipsum}

\usepackage{algpseudocode}

\newcommand{\gom}{\texttt{GoM}\xspace}

\newtcbox{\roundedtag}[1][]{
  on line,
  colback=tagbackground,    
  colframe=tagborder,       
  coltext=tagforeground,    
  boxrule=0.5pt,            
  arc=3pt,                  
  outer arc=3pt,
  top=-1.5pt,
  bottom=-1.5pt,
  left=-1pt,
  right=-1pt,
  #1
}

\newcommand{\roundedtagWithSymbol}[2]{%
  \noindent#1\;\textbf{\textit{#2}}\xspace
}

\newcommand{\quotes}[1]{``#1''}

\newcommand{\icon}[2][0.2in]{%
  \raisebox{-0.1\height}{\includegraphics[width=#1]{#2}}\xspace%
}
\newcommand{\rawImage}[1][0.12in]{%
  \icon[#1]{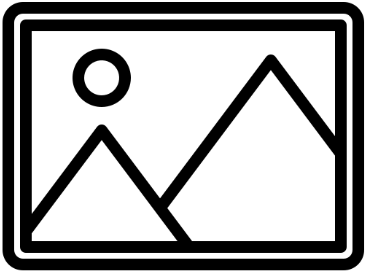}%
}
\newcommand{\seg}[1][0.12in]{%
  \icon[#1]{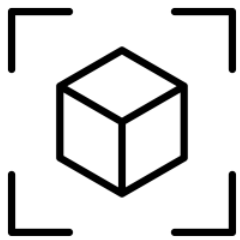}%
}
\newcommand{\segNumObj}[1][0.12in]{%
  \icon[#1]{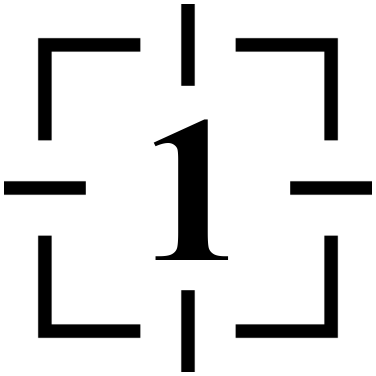}%
}
\newcommand{\gomNumObj}[1][0.12in]{%
  \icon[#1]{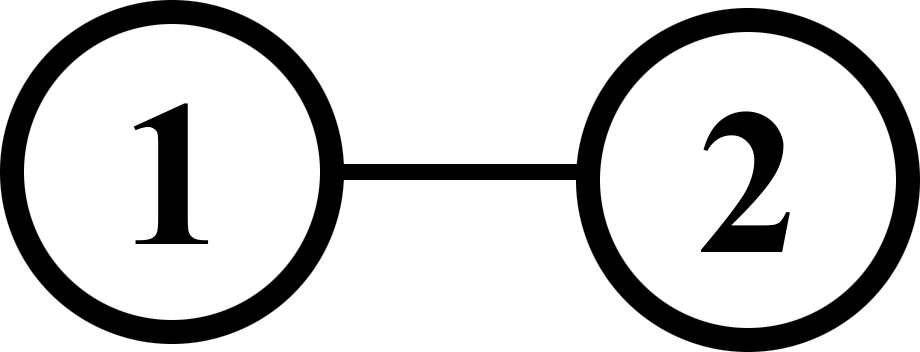}%
}
\newcommand{\gomLabObj}[1][0.12in]{%
  \icon[#1]{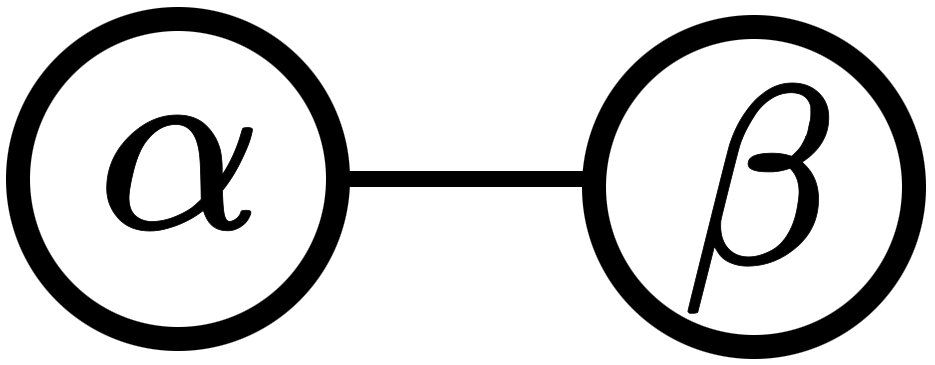}%
}
\newcommand{\gomNumObjLabRel}[1][0.12in]{%
  \icon[#1]{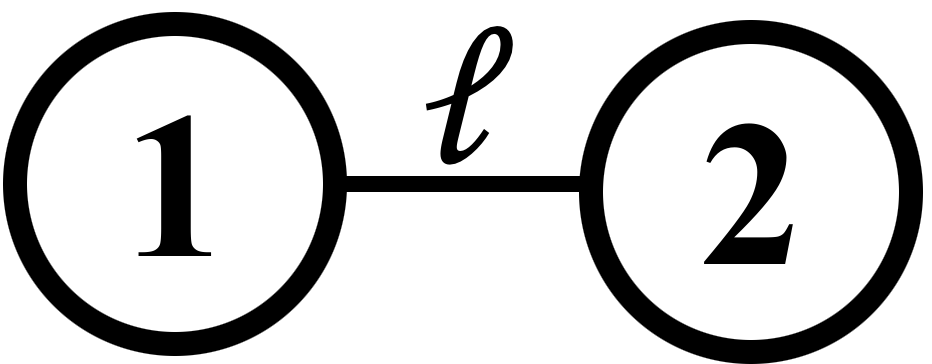}%
}
\newcommand{\gomLabObjLabRel}[1][0.12in]{%
  \icon[#1]{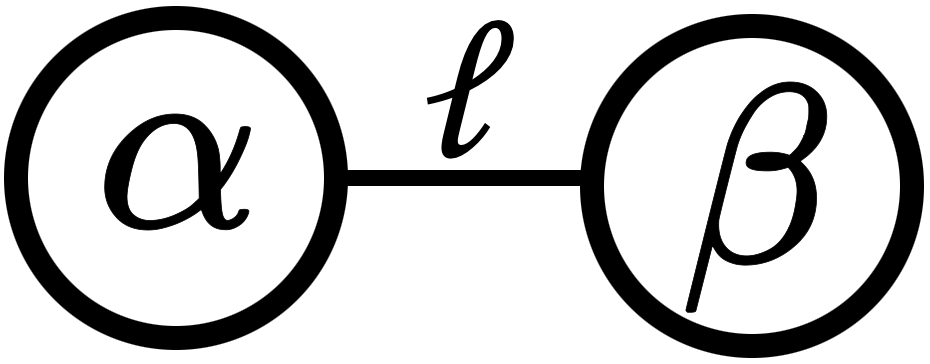}%
}

\definecolor{tagbackground}{HTML}{EBEBEB}
\definecolor{tagborder}{HTML}{888888}
\definecolor{tagforeground}{HTML}{000000}

\definecolor{tablehighlight}{HTML}{F3F3F3}
\definecolor{tablebordercolor}{HTML}{363636}
\definecolor{cellhighlight}{HTML}{E9E9E9}

\definecolor{plotbackground}{HTML}{F3F3F3}
\definecolor{gray}{HTML}{394144}
\definecolor{blue-1}{HTML}{1876D2} 
\definecolor{blue-2}{HTML}{60D0E2}
\definecolor{orange-1}{HTML}{FBA418} 
\definecolor{orange-2}{HTML}{FDCC7B}
\definecolor{purple-1}{HTML}{4D4DFF} 
\definecolor{purple-2}{HTML}{8F8FFF}
\definecolor{red-1}{HTML}{F00000} 
\definecolor{red-2}{HTML}{FF7575}

\definecolor{baseline}{HTML}{F2F2F2}
\definecolor{gom}{HTML}{F0F4FF}
\definecolor{error}{HTML}{D10000}
\definecolor{correct}{HTML}{0A9E4D}

\pgfdeclareverticalshading{bluegradient}{100bp}{
  color(0bp)=(blue-1);
  color(30bp)=(blue-1);
  color(60bp)=(blue-2);
  color(100bp)=(blue-2)
}
\pgfdeclareverticalshading{orangegradient}{100bp}{
  color(0bp)=(orange-1);
  color(30bp)=(orange-1);
  color(60bp)=(orange-2);
  color(100bp)=(orange-2)
}
\pgfdeclareverticalshading{purplegradient}{100bp}{
  color(0bp)=(purple-1);
  color(30bp)=(purple-1);
  color(60bp)=(purple-2);
  color(100bp)=(purple-2)
}
\pgfdeclareverticalshading{redgradient}{100bp}{
  color(0bp)=(red-1);
  color(30bp)=(red-1);
  color(60bp)=(red-2);
  color(100bp)=(red-2)
}